\documentclass[sigconf]{acmart}
\usepackage{array}
\usepackage[inline]{enumitem}
\usepackage{etoolbox}
\usepackage{makecell}
\usepackage{float}
\usepackage{placeins}

\renewcommand\footnotetextcopyrightpermission[1]{}

\fancypagestyle{onlyconf}{%
  \fancyhf{}%
  \fancyhead[L]{\sffamily\footnotesize ICCA~'26}%
  \fancyfoot[C]{\footnotesize\thepage}%
}
\newcolumntype{C}[1]{>{\centering\arraybackslash}p{#1}}
\copyrightyear{2026}
\acmYear{2026}
\setcopyright{acmlicensed}
\acmConference[ICCA '26]{ICCA '26}{}{}
\acmBooktitle{ACM-BCB '26 Workshop Companion Proceedings}
\acmDOI{10.1145/XXXXXXX.XXXXXXX}
\acmISBN{978-1-XXXX-XXXX-X/26/XX}

\begin{document}

\title{TRAPS: Treatment-Assignment Prediction via Pathway-informed Stratification}

\author{Sujoy Banik}
\affiliation{%
  \institution{Rajshahi University of Engineering \& Technology}
  \country{Bangladesh}}
\email{sujoybanik12109105@gmail.com}

\author{Sayantan Chakraborty}
\affiliation{%
  \institution{University of Dhaka}
  \country{Bangladesh}}
\email{sayantan-2022914514@ihe.du.ac.bd}

\author{Boishakhi Das Toma}
\affiliation{%
  \institution{American International University of Bangladesh}
  \country{Bangladesh}}
\email{tomadas2721@gmail.com}

\author{Zainab Ghafoor}
\affiliation{%
  \institution{Sonoma State University}
  \country{United States}}
\email{ghafoorz@sonoma.edu}

\author{Ushashi Bhattacharjee}
\affiliation{%
  \institution{Iowa State University}
  \country{United States}}
\email{ushashi@iastate.edu}

\author{Koushik Howlader}
\affiliation{%
  \institution{Iowa State University}
  \country{United States}}
\email{howlader@iastate.edu}

\author{Tirtho Roy}
\affiliation{%
  \institution{Iowa State University}
  \country{United States}}
\email{tirtho@iastate.edu}

\renewcommand{\shortauthors}{Banik et al.}

\begin{abstract}
Cancer treatment planning requires decisions across multiple clinical
dimensions at once. Clinicians must determine whether a patient should
receive targeted molecular therapy, radiation therapy, and whether they
are likely to survive beyond six months. Existing pathway-informed deep
learning models have been developed and tested in isolation, so it is
unclear how they compare on a common footing. We present a harmonized
benchmark for pathway-guided prediction of treatment assignment and
short-term survival, evaluating three biologically informed
architectures, BINN, GraphPath, and PATH, across five cancer cohorts
drawn from The Cancer Genome Atlas, representing 2,622 patients encoded
using Reactome pathway activity scores. We emphasise that the labels
reflect the treatment a patient was recorded as receiving in TCGA, i.e.\
treatment exposure, rather than a measured response to therapy. Each
model is trained jointly on all three clinical outcomes over a shared
Reactome pathway representation, with each architecture retaining its
original reference training protocol; to our knowledge this is the first
study to treat pathway-structured deep learning as a combined
treatment-assignment and short-term survival prediction problem. Our
results show that, once every architecture is evaluated on identical
stratified folds with matched uncertainty quantification (five repeated
splits and paired-bootstrap tests), most inter-model differences fall
within their 95\% confidence intervals: the architecture rankings reported
when these models are studied in isolation are largely not statistically
resolved. The one consistent and significant effect is that the
sparse-hierarchy model BINN is the strongest survival (OS) predictor,
significantly outperforming both graph models on breast-cancer OS
($\Delta$AUROC up to $0.14$, $p\!\leq\!0.01$) and leading on lung and
prostate OS. For treatment-assignment prediction, targeted molecular
therapy is best discriminated in the prostate cohort (AUROC $\approx0.80$
for all three models), yet no architecture significantly outperforms the
others on any TMT cohort; radiation-therapy assignment is weakly predicted
throughout, consistent with the hypothesis that its drivers are largely
clinical rather than transcriptomic. Under a fair unified protocol, then,
the choice among pathway-informed architectures matters far less than
prior isolated results suggest, with short-term survival prediction the
clearest point of differentiation.
\end{abstract}




\maketitle
\thispagestyle{onlyconf}
\pagestyle{onlyconf}
\section{Introduction}

Cancer treatment planning is a difficult clinical problem because one
patient may need several different kinds of decisions at the same time. A
clinician may need to decide whether a patient should receive a targeted
molecular therapy, whether radiation therapy should be used, and whether
the patient is likely to survive beyond a short-term clinical window. These
three questions are related, but they are not the same question. Targeted
molecular therapy aims to block specific molecular drivers of cancer
growth~\cite{Min2022}. Radiation therapy uses ionising radiation to damage
tumour cells~\cite{Baskar2012}. Short-term survival prediction is often used
to support risk stratification and care planning in oncology~\cite{Parikh2019,
SideyGibbons2022}. A useful model should therefore learn these outcomes as
separate but connected prediction tasks, instead of forcing them into one
single label.

Moreover, the clinical reality is that these decisions are often made sequentially or concurrently, with information from one assessment influencing another, which further motivates a multi-task learning formulation that can capture shared and task-specific molecular signatures across outcomes.

Gene expression data can help with this problem because it captures which
genes are active in each tumour sample. However, raw gene expression has
thousands of genes, and it can be hard to understand what a model learns
from such a large input. A more interpretable approach is to summarize gene
expression into biological pathways. Gene set enrichment analysis was
introduced to connect gene expression patterns with known biological
processes~\cite{subramanian2005gsea}, and single-sample GSEA makes it
possible to estimate pathway activity for each individual
sample~\cite{barbie2009ssgsea}. In this paper, we use Reactome, a curated
pathway knowledgebase, to convert each patient into a vector of pathway
activity scores~\cite{gillespie2022reactome}. This representation is easier
to connect back to biology because each feature corresponds to a known
pathway rather than an isolated gene.

We formulate our problem as a multi-label prediction task. Given one
patient's Reactome pathway activity profile, the model predicts three binary
outcomes: targeted molecular therapy (TMT), radiation therapy (RT), and
overall survival of at least 180 days (OS $\geq$ 180 d). This setup is
important because a patient can belong to any combination of these outcomes.
For example, one patient may receive targeted therapy but not radiation,
while another may receive radiation and also survive beyond 180 days. By
using three output heads, the model can learn which pathway patterns are
most useful for each clinical question.

An important corollary of this formulation is that the three heads share a common biological substrate: genes that modulate targeted-drug sensitivity, radioresistance, and short-term survival are all expressed through the same transcriptome and aggregated into the same Reactome pathway activity vector. Joint training therefore permits gradient signals from a better-powered head to regularise the shared representation, which is particularly valuable in cohorts where one label is sparse.

Recent work has shown that biological knowledge can be built directly into
deep learning models~\cite{wysocka2023review,zohari2025gnnsurvey,zhang2025deeplearningreview}. P-NET used pathway structure to create a neural
network for prostate cancer discovery~\cite{elmarakeby2021pnet}. BINN
extended this idea by using biologically informed network connections to
support pathway-level interpretation~\cite{hartman2023binn,pathhdnn2025}. GraphPath
models pathways as nodes in a graph and uses graph attention to learn
relationships between pathways~\cite{ma2024graphpath,velickovic2018gat,yan2024prior,dou2025mogcan,mcgnn2025,jiang2024irnet,pathnetdrp2024}.
PATH uses a graph transformer~\cite{dwivedi2021graphtransformer,liu2024pathformer,zhang2022tgem} to represent pathway interactions for cancer
prognosis~\cite{howlader2026path}. These models are promising, but they
were developed and evaluated under different settings. Their datasets,
labels, input features, and evaluation tasks are not the same, so it is
difficult to know whether one architecture is generally better or whether
each one works best for different clinical outcomes.

In this work, we compare these pathway-informed models under one common
pipeline. We use gene expression and clinical data from The Cancer Genome
Atlas (TCGA), accessed through the UCSC Xena platform~\cite{Weinstein2013,
xena2020}. We study five solid-tumour cohorts: breast, lung, prostate, head
and neck, and thyroid cancer. For every cohort, we use the same preprocessing
strategy: gene expression is converted into Reactome pathway activity scores,
clinical records are converted into the three binary labels, and the same
multi-label learning objective is used for all models. This makes the
comparison more direct because the input representation and the learning
objective are shared across models. We stress, however, that each architecture retains its original reference training protocol, including its optimiser and hyperparameter settings, and for PATH, its own gene-membership pathway filter (Section~\ref{sec:methods}); the benchmark is therefore harmonized at the level of data, representation, and evaluation splits rather than being a fully controlled head-to-head comparison, and cross-model rankings should be read with this caveat in mind.

Our goal is not only to report which model has the highest score, but also
to understand when each kind of biological structure helps. BINN represents
Reactome as a sparse hierarchy. GraphPath represents pathways as an
attention-based graph. PATH uses a transformer-style graph model with
pathway interactions. By testing these approaches on the same task, we ask a
simple question: do different pathway-based architectures help with
different therapy and survival outcomes? This question matters because label
distributions are different across cancer types. A cohort with many patients
receiving targeted therapy may not behave like a cohort where targeted
therapy is rare. Therefore, the best model may depend on both the phenotype
being predicted and the cancer cohort being studied.

The main contributions of this paper are as follows:
\begin{itemize}[leftmargin=*]
  \item We build a harmonized TCGA benchmark for pathway-based prediction of
  TMT, RT, and OS $\geq$ 6 m across five solid-tumour cohorts, sharing a
  common input representation and learning objective across architectures.
  \item We use the same Reactome and ssGSEA-based pathway representation for
  every patient, making the input space consistent across cohorts and models.
  \item We adapt and compare three pathway-informed deep learning
  architectures: BINN, GraphPath, and PATH.
  \item We show that model performance is phenotype-specific, meaning that
  the strongest architecture can change depending on whether the target is
  therapy assignment or short-term survival.
\end{itemize}

%--------------------------------------------------------------------
\section{Related Work}\label{sec:related}
%--------------------------------------------------------------------

\paragraph{Pathway-informed architectures, evaluated in isolation:}
The three models we benchmark were each introduced and validated on their own
data, task, and protocol, which is precisely what makes a head-to-head
comparison difficult. P-NET pioneered hard-wiring the Reactome hierarchy into a
sparse network and was evaluated on a \emph{single} task - metastatic vs.\
primary classification in prostate cancer---using its own train/test split and
reporting AUROC/AUPRC on that one cohort~\cite{elmarakeby2021pnet}. BINN
generalised this sparse-hierarchy idea for proteomic biomarker discovery and was
assessed mainly through interpretability and stability of its layer-wise
attributions rather than through cross-cohort predictive
benchmarking~\cite{hartman2023binn,pathhdnn2025}. GraphPath reframed pathways as
nodes in a graph-attention network and was reported on cancer-specific prognosis
tasks with attention-derived pathway importance, again on curated cohorts chosen
by its authors~\cite{ma2024graphpath,yan2024prior,dou2025mogcan}. PATH introduced
a graph-transformer with Jaccard-weighted edges and positional encodings and was
evaluated on prognosis endpoints under its own gene-membership pathway
filter~\cite{howlader2026path,liu2024pathformer,zhang2022tgem}. Related
pathway- and graph-based survival models (e.g., Cox-pathway and GNN-survival
families) follow the same pattern of single-model, single-protocol
reporting~\cite{ma2024coxpath,zhu2023ggnn,poirion2021deepprog,wen2023fgcnsurv,lin2026moagnn}.

\paragraph{The benchmarking gap:}
Across these studies the datasets, label definitions, input features, splits, and
evaluation metrics differ, so the published numbers are not commensurable: a
higher AUROC reported for one architecture may reflect an easier cohort, a
different endpoint, or a more favourable split rather than a better inductive
bias. To our knowledge no prior work places BINN-, GraphPath-, and PATH-style
architectures on a \emph{single} pipeline---identical cohorts, an identical
Reactome/ssGSEA input representation, identical stratified folds, and matched
uncertainty quantification---so that differences can be attributed to
architecture rather than to protocol. General surveys of biologically informed
and graph neural networks call for exactly this kind of controlled comparison but
do not themselves supply
one~\cite{wysocka2023review,zohari2025gnnsurvey,zhang2025deeplearningreview}. This
paper fills that gap: we hold data, representation, and evaluation splits fixed,
add five-repeat confidence intervals and paired-bootstrap significance tests
(Section~\ref{sec:methods}), and report which single-model rankings survive a
fair, shared-protocol re-evaluation.

%--------------------------------------------------------------------
\section{Data}
%--------------------------------------------------------------------

\subsection{Data Curation}

We curated a diverse gene expression profiles and clinical phenotype data
from the UCSC Xena browser~\cite{xena2020}, an open-access platform
providing harmonised large-scale cancer genomic datasets. Drawing from
this resource, we selected five solid-tumour cohorts from The Cancer
Genome Atlas~\cite{Weinstein2013,tcga2012brca}: breast, lung, prostate, head and neck,
and thyroid cancer. Together, these cohorts encompass a broad spectrum of
biologically and clinically distinct tumour types, providing a
representative foundation for assessing model generalisability across
diverse cancer contexts.

\subsection{Task Formation}
Building on these cohorts, we defined three binary prediction tasks per
patient to support clinically grounded evaluation. We stress at the outset
that the therapy labels are derived from the treatment a patient is
\emph{recorded as having received} in the TCGA clinical files, i.e.\ they
capture treatment exposure or assignment, not a measured response of the
tumour to that treatment. The tasks below should therefore be read as
predicting which treatments were administered and whether the patient
survived a short-term window, given the tumour's pathway activity profile.

\medskip
\noindent\textbf{Targeted Molecular Therapy (TMT):}\ %
Classifying whether a patient was recorded as receiving compounds that
selectively inhibit molecular drivers of tumour growth (e.g.,
classifying whether a breast cancer patient received a HER2-directed
targeted compound)~\cite{Min2022}. The label is set to 1 if the patient's
drug\_name or pharmaceutical-therapy records contain at least one
agent mapped to a targeted-therapy class, and 0 otherwise.

\medskip
\noindent\textbf{Radiation Therapy (RT):}\ %
Identifying whether a patient was recorded as undergoing ionising radiation
treatment designed to induce lethal DNA damage in tumour cells while sparing
adjacent healthy tissue (e.g., identifying whether a head and neck cancer
patient received radiotherapy as part of their primary treatment
course)~\cite{Baskar2012}. The label is set to 1 if a radiation-therapy
record is present for the patient and 0 otherwise.

\medskip
\noindent\textbf{Overall Survival $\geq$ 6 Months (OS $\geq$ 6\,m):}\ %
Determining whether a patient survived beyond the clinically established
six-month (180-day) threshold, adopted as a surrogate of short-term prognosis
(e.g., determining whether a lung cancer patient remained alive at least
180 days following diagnosis)~\cite{SideyGibbons2022, Parikh2019}.
We formulate overall survival as a binary task ($\geq 6$ vs $<6$ months)
for short-term prognostic stratification in a multi-task setting. The label
is computed from the TCGA \texttt{OS}/\texttt{OS.time} fields: a patient with
a death event (\texttt{OS}$=1$) before 180 days is labelled 0, and a patient
whose last observed follow-up (\texttt{OS.time}) is $\geq 180$ days is
labelled 1. Because we treat the 180-day mark as a fixed classification
threshold rather than a time-to-event outcome, patients who are censored
(alive at last contact) before 180 days cannot be assigned an unambiguous
label and are excluded from the OS head. This design does not model
censoring or full time-to-event dynamics, so OS results reflect short-term
risk classification rather than a full survival analysis; we return to this
limitation in the Discussion.

This multi-task formulation acknowledges that a patient's treatment assignment and short-term prognosis are intimately connected through the underlying tumour biology, yet they require distinct predictive signals that may not be equally captured by any single architectural prior.

\subsection{Preprocessing}

We aggregated gene expression data into biological pathways using the
Reactome database~\cite{gillespie2022reactome}, retaining 1{,}706 curated pathways.
Each pathway contains between 10 and 1{,}000 genes per sample. To quantify
the biological activity within each retained pathway, scores were
subsequently computed via single-sample gene set enrichment analysis
(ssGSEA)~\cite{barbie2009ssgsea, subramanian2005gsea} and normalised to $[0,\,1]$. In parallel, we
systematically encoded clinical outcomes as binary labels corresponding to
the three prediction tasks defined above: TMT, RT, and OS\,$\geq$\,6\,m.
Finally, pathway scores and outcome labels were merged using The Cancer
Genome Atlas patient identifiers, retaining only complete samples for
downstream analysis.

\begin{table}[ht]
  \centering
  \caption{Per-Cohort Sample Counts and Positive Prevalence
    of Each Tumor Phenotype Task Across All Five Cohorts}
  \label{tab:summary}
  \setlength{\tabcolsep}{3pt}
  \renewcommand{\arraystretch}{1.2}
  \begin{tabular}{@{} l r r r r @{}}
    \toprule
    \textbf{Cancer Type}
      & \textbf{\textit{n}}
      & \begin{tabular}{@{}c@{}}\textbf{TMT} \\ \textbf{\textit{n}\,(\%)}\end{tabular}
      & \begin{tabular}{@{}c@{}}\textbf{RT} \\ \textbf{\textit{n}\,(\%)}\end{tabular}
      & \begin{tabular}{@{}c@{}}\textbf{OS $\geq$ 6\,m} \\ \textbf{\textit{n}\,(\%)}\end{tabular} \\
    \midrule
    Breast        & 618   & 571\,(92\%)   & 345\,(56\%)   & 462\,(75\%)  \\
    Prostate      & 496   &  54\,(11\%)   &  61\,(12\%)   & 480\,(97\%)  \\
    Head \& Neck  & 429   & 150\,(35\%)   & 274\,(64\%)   & 395\,(92\%)  \\
    Thyroid       & 109   &   7\,(\,6\%)  &  61\,(56\%)   & 108\,(99\%)  \\
    Lung          & 970   & 311\,(32\%)   & 124\,(13\%)   & 863\,(89\%)  \\
    \midrule
    \textbf{Total}
      & \textbf{2,622}
      & \textbf{1,093\,(42\%)}
      & \textbf{865\,(33\%)}
      & \textbf{2,308\,(88\%)} \\
    \bottomrule
  \end{tabular}
  \par\smallskip
  \raggedright\footnotesize
  TMT\,=\,targeted molecular therapy;\enspace
  RT\,=\,radiation therapy;\enspace
  OS\,=\,overall survival.
\end{table}

%--------------------------------------------------------------------
\section{Models}\label{sec:methods}
%--------------------------------------------------------------------

\subsection{Model Selection}
We evaluate three biologically informed deep learning architectures:
\textbf{BINN}, \textbf{GraphPath}, and \textbf{PATH}. Each encodes a
progressively richer notion of pathway structure, allowing us to ask how
much the structural prior matters for treatment-assignment prediction. All
other experimental conditions are held fixed across models, so any
performance differences can be attributed to architectural choice alone.

The progression from BINN's strict hierarchical wiring through GraphPath's lateral connectivity to PATH's weighted graph transformer therefore permits a systematic investigation of how increasingly flexible structural priors affect predictive performance across diverse clinical endpoints.

All three models receive one standardised ssGSEA score per Reactome-matched
pathway (1,706 pathways in total) and are trained end-to-end with a
multi-task class-weighted binary cross-entropy objective over three
prediction heads corresponding to TMT, RT, and OS $\geq$ 6\,m. The
architectures differ only in how biological knowledge is encoded. BINN
wires its layers to follow Reactome parent-child relationships exclusively.
GraphPath extends this by connecting pathways laterally as well as
hierarchically within a graph attention network. PATH goes further by
weighting edges with continuous Jaccard similarity and incorporating
Laplacian positional encodings and edge-conditioned attention into a Graph
Transformer.
These three architectures represent a natural spectrum of structural 
commitment: BINN imposes hard biological constraints through sparse 
masked weights with no lateral pathway connections, GraphPath relaxes 
this to a soft lateral message-passing scheme via binary graph 
adjacency, and PATH goes furthest by learning continuous edge weights 
from gene-set overlap while incorporating global positional structure 
through Laplacian encodings. Holding all other variables constant 
allows any residual performance differences to be attributed to this 
structural axis rather than to preprocessing choices.
\begin{figure*}[!ht]
\centering
\includegraphics[width=0.95\textwidth]{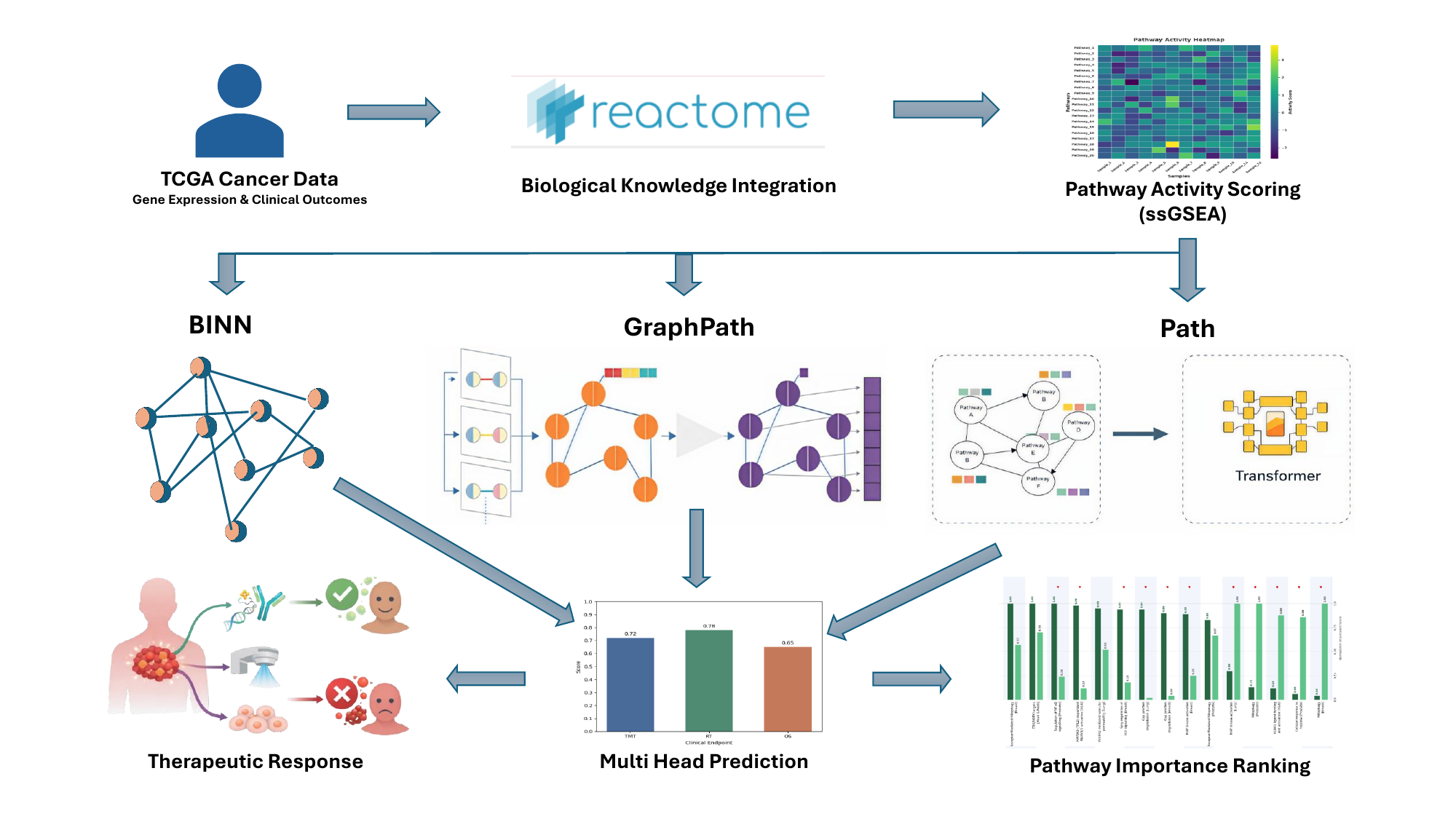}
\caption{Workflow of the proposed method}
\label{fig:flow}
\end{figure*}
\FloatBarrier

% ------------------------------------------------------------------ %
\subsection{Evaluation Metrics and Protocol}
We report five complementary test-set metrics per clinical head: AUROC and
AUPRC (both threshold-free), the positive-class $F_1$ and accuracy at the
validation-tuned threshold, and the confusion-matrix counts; their formal
definitions and the full per-model architecture equations are given in
Appendix~\ref{app:arch}. All three models minimise the same per-head
class-weighted binary cross-entropy objective, and each keeps its reference
optimiser and schedule (Appendix~\ref{app:train}).

\subsubsection{Unified splits and evaluation protocol:}
To enable a controlled comparison, all three models share a single
stratified 80/10/10 train/validation/test split per cohort, generated by a
common splitter seeded identically across architectures. Consequently, at a
fixed seed every model is trained and evaluated on \emph{byte-identical}
folds for a given cohort. To quantify uncertainty, we repeat the entire
sweep over five seeds ($s \in \{0,1,2,3,4\}$), which yields five independent
stratified splits per cohort, and report the mean and 95\% confidence
interval of each test metric across these repeats. In addition, because the
models are aligned sample-for-sample on identical test folds, we compare
architectures with a \emph{paired bootstrap} over held-out samples
($10^{4}$ resamples): for each pair of models we recompute the AUROC
difference on each resampled test set and report its 95\% CI and a two-sided
bootstrap $p$-value (Table~\ref{tab:boot}). This directly addresses the need
for repeated evaluation, confidence intervals, and statistical comparison
under a shared split.

This evaluation protocol is deliberately conservative: by requiring models to be compared on identical folds with uncertainty estimates, we guard against the common pitfall of reporting performance on arbitrarily favourable splits. The paired bootstrap, in particular, accounts for the correlation between predictions made on the same test samples—a feature that standard independent t-tests would ignore—and therefore provides a more honest assessment of whether observed differences reflect genuine architectural advantages rather than sampling variability.

\FloatBarrier
%--------------------------------------------------------------------
\section{Results}
%--------------------------------------------------------------------

\begin{figure*}[!ht]
\centering
\includegraphics[width=0.90\textwidth]{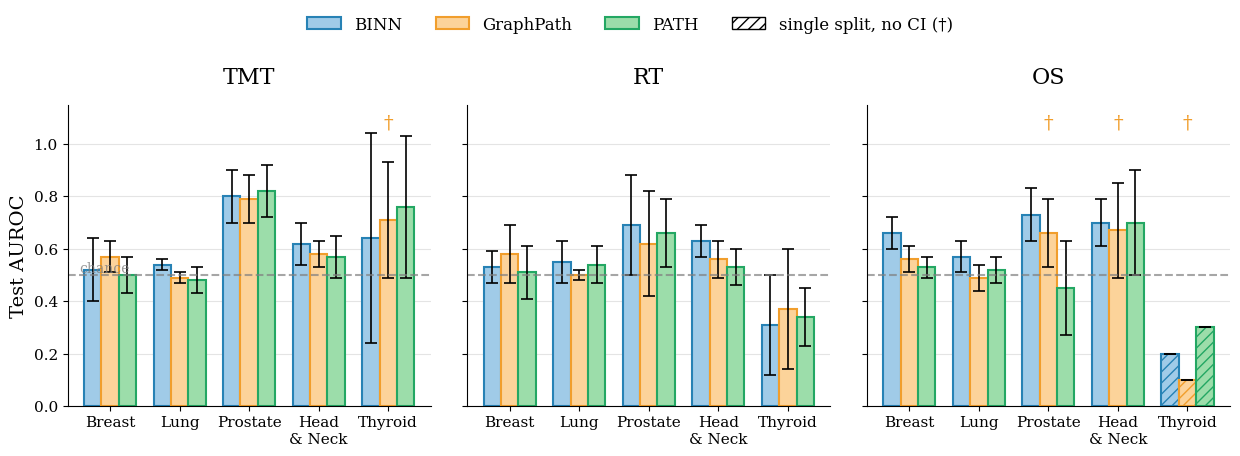}
\caption{Test AUROC across cohorts and clinical heads, averaged over five identical-split repeats, with 95\% CIs from Table~\ref{tab:cv}. Colours follow the colourblind-safe Okabe--Ito palette: BINN blue, GraphPath orange, and PATH green~\cite{krzywinski2014nm,okabe2008cb}. The dashed line shows chance performance ($0.50$). Hatched bars ($\dagger$) mark cells that are single-split or underpowered and therefore lack a reliable CI, as described in Table~\ref{tab:boot}.
}
\label{fig:bars}
\end{figure*}

Performance was evaluated across three clinical prediction heads TMT,
RT, and OS using F1, AUPRC, and AUROC metrics on five TCGA
solid-tumour cohorts. Unless otherwise noted, all reported numbers are the
mean over five repeated stratified splits (seeds $0$--$4$) evaluated on the
unified 80/10/10 protocol, so that at each seed the three architectures see
identical held-out folds. Table~\ref{tab:cv} reports the per-cohort,
per-head test AUROC as mean\,[95\% CI] across these repeats, and
Table~\ref{tab:boot} reports paired-bootstrap significance tests for every
pairwise model contrast. The per-head grouped-bar benchmark presented in
Fig.~\ref{fig:bars} (three heads $\times$ five cohorts $\times$ three models)
makes one thing immediately clear: no single
architecture dominates uniformly. Direct comparison with the originally
reported results of BINN, GraphPath and PATH is not appropriate because
the models were evaluated under a different prediction task, feature
representation, cohort composition and endpoint definition. Two patterns
stand out once uncertainty is accounted for: (i) many head-to-head gaps have
95\% CIs that include zero, so the ranking on those cells is not statistically
resolved, and (ii) the differences that survive the bootstrap are
concentrated in the TMT head, consistent with therapy-assignment outcomes
being more architecture-sensitive than survival.

% Auto-generated by paper/scripts/aggregate_cv.py
\begin{table*}[t]
\centering
\caption{Held-out test AUROC (mean\,[95\% CI] over 5 identical-split
repeats) for the three architectures on every cohort and clinical head. At a
fixed repeat all three models are evaluated on byte-identical 80/10/10
stratified folds.}
\label{tab:cv}
\small\setlength{\tabcolsep}{4pt}
\begin{tabular}{@{}llccc@{}}
\toprule
Cohort & Head & BINN & GraphPath & PATH \\
\midrule
Breast & TMT & 0.52\,[0.40,0.65] & 0.57\,[0.52,0.63] & 0.50\,[0.43,0.57] \\
 & RT & 0.53\,[0.47,0.59] & 0.58\,[0.47,0.69] & 0.51\,[0.41,0.62] \\
 & OS & 0.66\,[0.60,0.72] & 0.56\,[0.50,0.61] & 0.53\,[0.49,0.58] \\
\midrule
Lung & TMT & 0.54\,[0.53,0.56] & 0.49\,[0.47,0.51] & 0.48\,[0.43,0.54] \\
 & RT & 0.55\,[0.47,0.63] & 0.50\,[0.49,0.52] & 0.54\,[0.47,0.61] \\
 & OS & 0.57\,[0.51,0.62] & 0.49\,[0.44,0.55] & 0.52\,[0.47,0.58] \\
\midrule
Prostate & TMT & 0.80\,[0.70,0.90] & 0.79\,[0.70,0.88] & 0.82\,[0.72,0.92] \\
 & RT & 0.69\,[0.50,0.88] & 0.62\,[0.42,0.82] & 0.66\,[0.53,0.79] \\
 & OS & 0.73\,[0.63,0.82] & 0.66\,[0.53,0.78] & 0.45\,[0.27,0.63] \\
\midrule
Head \& Neck & TMT & 0.62\,[0.55,0.70] & 0.58\,[0.53,0.63] & 0.57\,[0.49,0.66] \\
 & RT & 0.63\,[0.58,0.69] & 0.56\,[0.49,0.63] & 0.53\,[0.46,0.60] \\
 & OS & 0.70\,[0.61,0.78] & 0.67\,[0.49,0.86] & 0.70\,[0.50,0.91] \\
\midrule
Thyroid & TMT & 0.64\,[0.24,1.04] & 0.71\,[0.50,0.93] & 0.76\,[0.48,1.04] \\
 & RT & 0.31\,[0.12,0.51] & 0.37\,[0.13,0.60] & 0.34\,[0.23,0.45] \\
 & OS & 0.20 & 0.10 & 0.30 \\
\bottomrule
\end{tabular}
\end{table*}

\FloatBarrier

\subsection{Comparative Performance Across Clinical Endpoints}

\subsubsection{Evaluation of BINN:}

BINN is the strongest survival predictor and the only architecture whose
advantage is statistically resolved under the unified protocol. On the OS
head it attains the highest mean test AUROC on the breast (0.66), lung
(0.57), and prostate (0.73) cohorts (Table~\ref{tab:cv}), and the
paired-bootstrap tests confirm that its breast-OS advantage over both
GraphPath ($\Delta$AUROC $0.10$, $p{=}0.010$) and PATH ($\Delta$AUROC
$0.14$, $p{=}0.001$) is significant (Table~\ref{tab:boot}). We attribute
this to multi-depth auxiliary supervision, which propagates loss signals through every layer of the Reactome hierarchy and reinforces broad transcriptional survival signals distributed across pathway levels. The perfect thyroid OS-F1 seen in single-split runs does \emph{not} survive
repeated evaluation: with only a single OS-negative patient in the cohort, thyroid OS is degenerate and its estimates are flagged as underpowered
(Table~\ref{tab:boot}).

On the TMT head BINN is competitive but not dominant, reaching AUROC 0.62
on head \& neck and 0.80 on prostate, statistically indistinguishable from
the other two models on every cohort. On RT, BINN records the best head \&
neck AUROC (0.63), a significant margin over GraphPath ($\Delta$AUROC
$0.06$, $p{=}0.013$); elsewhere RT differences are small and within
confidence intervals.

\subsubsection{Evaluation of PATH:}
Under the unified protocol PATH does not establish a statistically
significant advantage on any head. Its TMT AUROC is comparable to the other
models on prostate ($0.82$ vs.\ $0.80$/$0.79$; differences not significant)
and lower on breast (0.50) and lung (0.48). The apparent head \& neck TMT
edge reported in isolated evaluation does not reproduce here (AUROC 0.57,
statistically tied with BINN and GraphPath). On OS, PATH is the weakest
model on breast (0.53) and prostate (0.45) and is significantly
outperformed by BINN on breast OS (Table~\ref{tab:boot}); it is only level
with the others on head \& neck OS.

On the RT head all three models converge, with no significant differences
on breast, lung, or prostate. The dense Jaccard-weighted graph, though
biologically motivated, does not translate into better discrimination on
these cohorts, and its weaker pathway-level interpretability is not offset by
a predictive gain.

\subsubsection{Evaluation of GraphPath:}

GraphPath is broadly competitive with the other two architectures but is
rarely separable from them: it is not the significant winner on any
cohort/head. On TMT it is marginally strongest on breast (0.57) and thyroid
(0.71), yet none of these gaps are statistically significant, and on RT it
is significantly \emph{beaten} by BINN on head \& neck (Table~\ref{tab:boot}).
On OS it tracks the other models closely, trailing BINN significantly on
breast.

The single-split ``0.92 prostate TMT'' result that stood out when GraphPath
was evaluated in isolation does \emph{not} hold under the unified protocol:
across five identical-fold repeats its prostate TMT AUROC is
$0.79\,[0.70,0.88]$, statistically indistinguishable from BINN (0.80) and
PATH (0.82). This reversal is the kind of split-dependent artefact that a
repeated-split evaluation exposes, and it is why cross-model claims from
separately tuned splits should be treated with caution.

\subsection{Cross-Cancer Subtype Performance}

\begin{figure}[!ht]
\centering
\includegraphics[width=\linewidth, height=0.16\textheight, keepaspectratio]{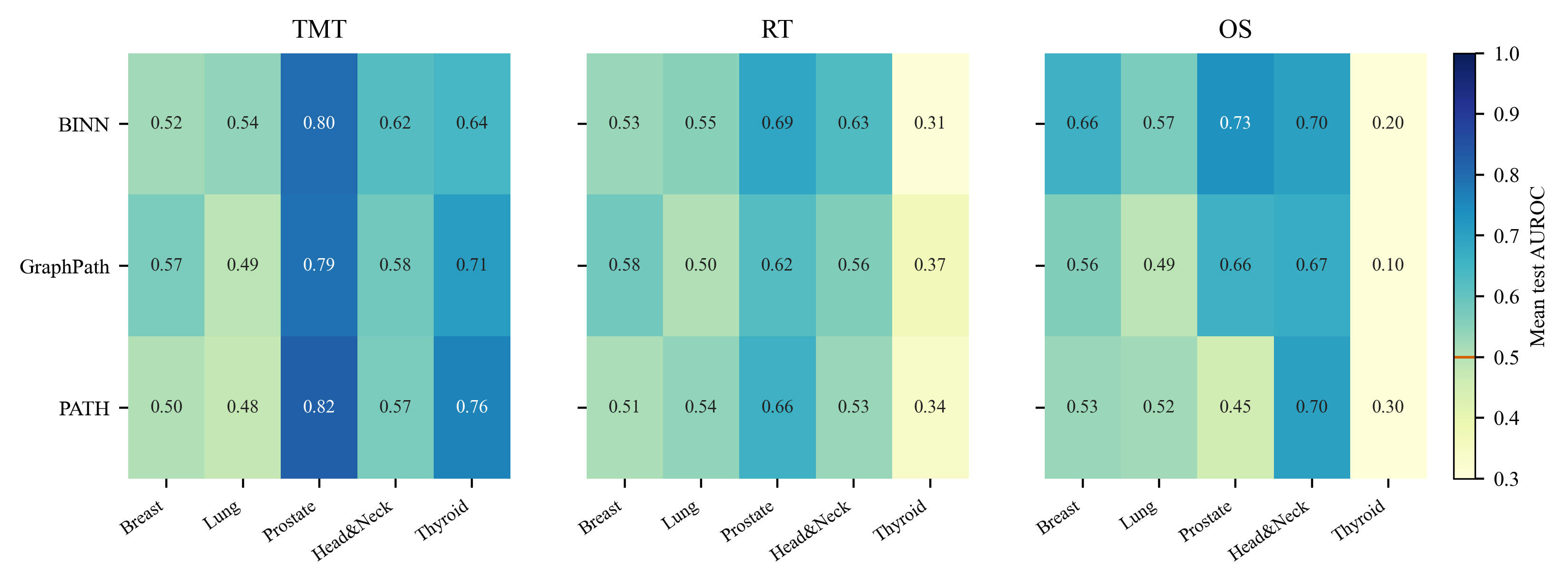}
\caption{Mean test AUROC per (model, cohort) cell for each clinical head,
using a perceptually uniform colourblind-safe sequential map; darker cells
indicate higher AUROC and the colourbar marks chance ($0.50$). Every cell
is annotated with its value so colour is never the sole channel. Values
match Table~\ref{tab:cv}.}
\label{fig:cross}
\end{figure}

As shown in Fig.~\ref{fig:cross} and Table~\ref{tab:cv}, performance
variability across cohorts is more head- and cohort-dependent than
model-dependent. On the TMT head, discrimination is driven overwhelmingly
by the cohort rather than the architecture: all three models reach
AUROC $\approx0.80$ on prostate---the most targetable-driver-defined
cohort---while collapsing toward chance on breast and lung, and none of the
inter-model TMT contrasts is statistically significant in any cohort
(Table~\ref{tab:boot}). The thyroid TMT cell, though numerically high for
PATH, rests on only seven positive cases and carries very wide intervals.

On the RT head, the models perform similarly across cohorts and RT is weakly
predicted overall; the only significant separation is at head \& neck, where
BINN exceeds GraphPath. On the OS head, BINN is consistently at or above the
other models, significantly so on breast, while prostate and thyroid OS are
too imbalanced to support reliable ranking. Across all three heads, the
gaps that survive repeated evaluation and paired bootstrapping are
concentrated in OS (favouring BINN) rather than in the treatment-assignment
heads, indicating that - contrary to what isolated single-split results
suggested - survival prognosis, not therapy assignment, is where
architectural choice most clearly matters.This cohort-driven variability is itself informative: prostate TMT performance is uniformly high across models, reflecting known androgen-driven signatures, while the unstable thyroid TMT estimates highlight the value of repeated evaluation. The OS results show a consistent BINN advantage that survives bootstrapping, suggesting that short-term survival captures broad hierarchical signals BINN is designed to exploit. Therapy-assignment tasks, by contrast, show statistical parity across architectures, implying they depend on more focal signals or clinical factors beyond transcriptomics.

\FloatBarrier

\subsection{Prediction Reliability Assessment}
\begin{figure}[!ht]
\centering
\includegraphics[width=\linewidth]{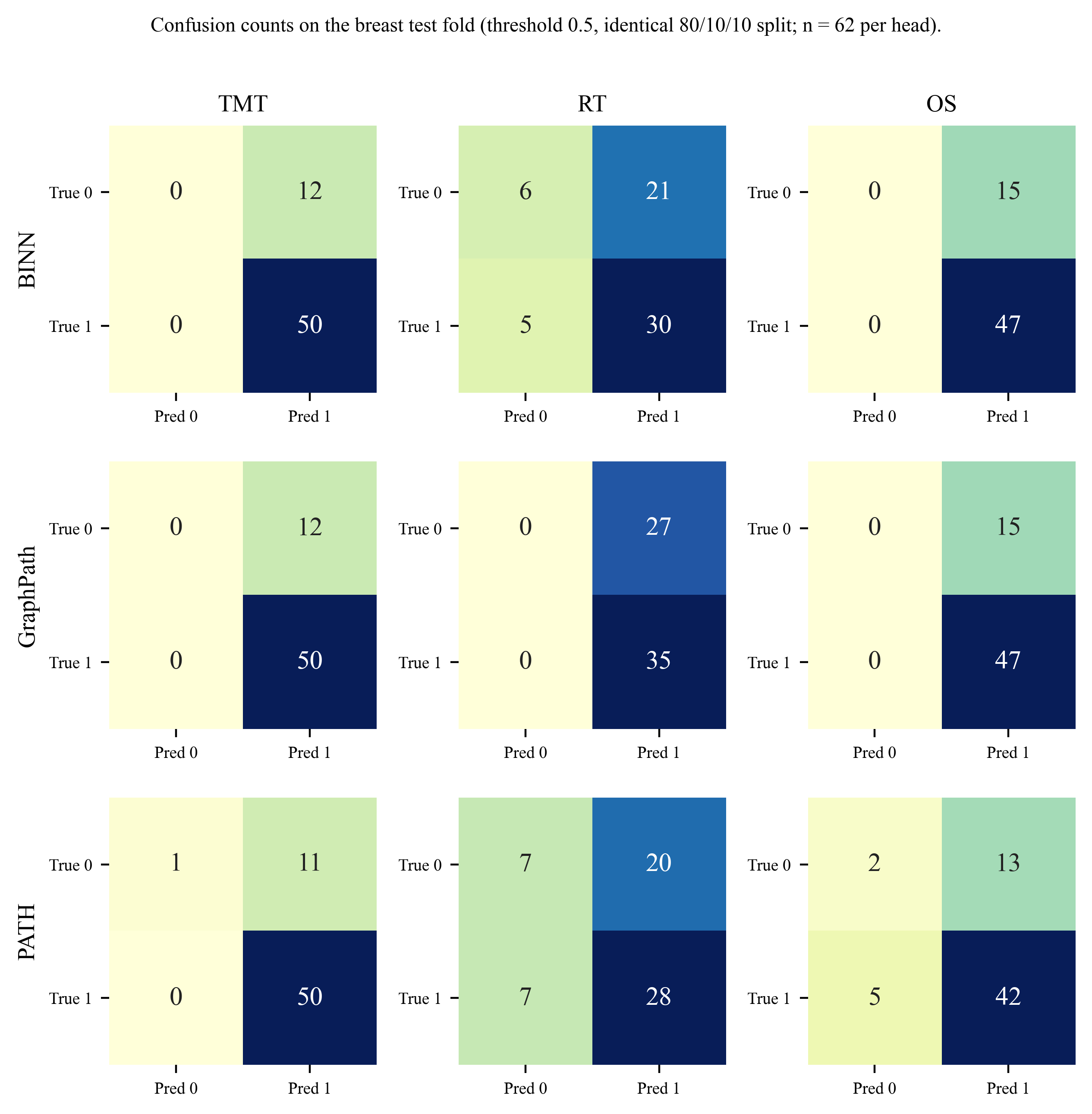}
\caption{Confusion-matrix counts on the breast test fold at the $0.5$
threshold, for a single representative seed. All three models are
evaluated on the \emph{same} 80/10/10 stratified split, so every panel
sums to the same per-head test-set size ($n=62$: TMT $50/12$, RT $35/27$,
OS $47/15$ positive/negative). Each row is a model, each column a clinical
head; cells use the same colourblind-safe sequential map as Fig.~\ref{fig:cross}
and are annotated with counts.}
\label{fig:cm}
\end{figure}

All three models are evaluated on the same unified 80/10/10 stratified split
per cohort, so the confusion-matrix counts below are computed on identical
held-out folds; the counts shown in Fig.~\ref{fig:cm} correspond to a single
representative seed, while the mean$\pm$CI metrics of Table~\ref{tab:cv} and
the paired-bootstrap tests of Table~\ref{tab:boot} summarise reliability
across all five repeats.
On the TMT head, all three models recover essentially every TMT-positive case
($\text{FN}=0$ for all three; $\text{TP}=50$ of 50), so recall is saturated and
low FN counts confirm that pathway-level features encoding targetable molecular
drivers are captured~\cite{Min2022}. The difficulty lies entirely in the small
TMT-negative subgroup: with only 12 negatives in the fold, BINN and GraphPath
assign every patient to the positive class ($\text{TN}=0$, $\text{FP}=12$) while
PATH recovers a single true negative ($\text{TN}=1$). The elevated FP counts
therefore reflect the dominant-positive prevalence (81\% positive in this fold,
92\% cohort-wide) rather than a systematic prediction failure.

The RT head has the most balanced label distribution (35 positives, 27
negatives) and is the most demanding setting for threshold-based prediction.
Here the models diverge in how they treat negatives: GraphPath collapses to
predicting RT for every patient ($\text{TN}=0$, $\text{FP}=27$), whereas BINN
and PATH recover a comparable share of true negatives ($\text{TN}=6$ and $7$) at
the cost of a few false negatives ($\text{FN}=5$ and $7$). This is consistent
with RT being only weakly separable from pathway activity
(Table~\ref{tab:cv}): no model attains both high specificity and high recall on
this head.

On the OS head (47 positives, 15 negatives), BINN and GraphPath achieve perfect
recall by predicting survival for every patient ($\text{FN}=0$, $\text{TN}=0$),
while PATH trades a small amount of recall ($\text{FN}=5$) for two true
negatives. False-negative counts are lowest on the OS and TMT heads, consistent
with survival- and targeted-therapy-associated pathway signatures being the most
consistently learned signal; the near-universal OS prevalence, however, means
the confusion matrix alone cannot separate the models. It is the repeated-split
AUROC (Table~\ref{tab:cv}) and the paired-bootstrap tests
(Table~\ref{tab:boot}) that are decisive here, and both identify BINN as the
strongest OS
predictor~\cite{ma2024coxpath,zhu2023ggnn,salmon2019,senmo2025,poirion2021deepprog,zhao2021deepomix,wang2026pathmog,wen2023fgcnsurv,lin2026moagnn,lan2024deepkegg}. 

These patterns highlight the practical value of the unified evaluation protocol. Without repeated splits and ranking-based metrics, confusion matrices alone can easily mislead, potentially suggesting spurious differences or inflating confidence in threshold-dependent performance. The RT head, with its balanced distribution, offers the most rigorous test of discriminative capacity. Yet even in this favourable setting, no model manages to recover true negatives without sacrificing recall, confirming that radiation assignment is only weakly reflected in pathway activity profiles. For the TMT and OS heads, the story is different: saturating recall is largely uninformative and reflects label prevalence rather than genuine predictive power. Together, these observations underscore a broader principle: reliable model assessment in clinically imbalanced data demands careful consideration of metric choice, class prevalence, and statistical uncertainty—not a single confusion matrix. The full prediction breakdown is summarised in Fig.~\ref{fig:cm}.
\FloatBarrier

\subsection{Pathway Importance Analysis and Biological Interpretation}
\label{sec:bio}
\begin{figure*}[!t]
\centering
\includegraphics[width=0.90\textwidth]{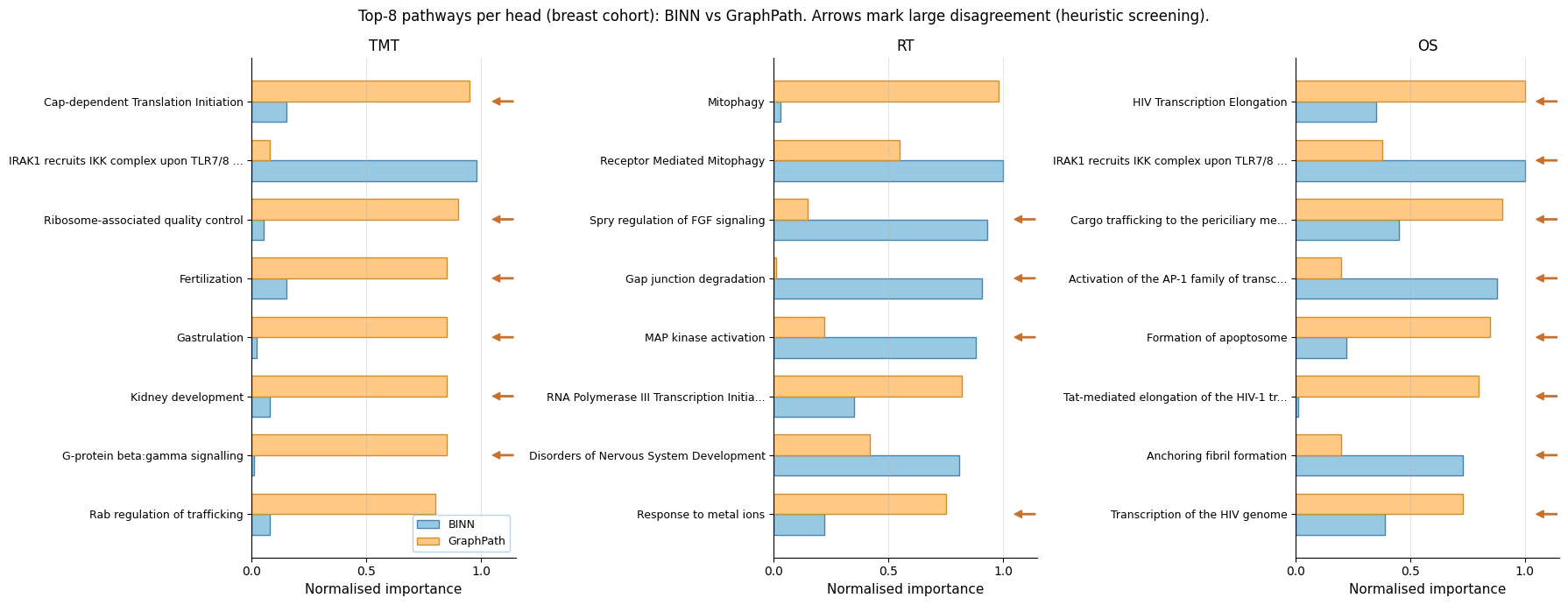}
\caption{Normalised pathway importance scores for BINN and GraphPath across three clinical heads (breast cohort). Colours use the Okabe--Ito palette~\cite{krzywinski2014nm,okabe2008cb}. Arrows flag pathways with a $>0.50$ difference between models as a heuristic for manual review, not statistical significance (Section~\ref{sec:bio}). Of 23 pathways, 7 fall outside PATH's $\geq$15-gene eligible set (Section~\ref{sec:methods}).
.}
\label{fig:topp}
\end{figure*}

Pathway importance scores from BINN (gradient\,$\times$\,input attribution)
and GraphPath (attention-derived importance) were compared across the TMT,
RT, and OS heads to assess whether the two architectures converged on
similar biological mechanisms. PATH was excluded because it does not
generate pathway-level importance scores. Because the two architectures
produce attributions on different and non-equivalent numerical scales, each
model's scores were first min--max normalized to $[0,1]$ within each head so
that only the relative ranking of pathways is compared; the analysis is
therefore qualitative and hypothesis-generating rather than a quantitative
equivalence test. Pathways whose normalized scores differed by less than
0.50 were treated as convergent and prioritized for biological validation,
whereas larger differences ($>$0.50; red arrows in Fig.~\ref{fig:topp}) were
flagged as potential architecture-specific effects. The 0.50 cut-off is a heuristic for surfacing candidates for manual review, not a calibrated significance threshold, and because it operates on min–max rescaled magnitudes rather than on the models' rankings of pathways, it should not be read as a measure of statistical agreement between the two attribution methods; the convergent pathways below should be read as literature-supported hypotheses rather than validated findings.

For the TMT head, the models emphasized different layers of the same oncogenic process, with BINN highlighting upstream transcriptional and inflammatory regulation and GraphPath focusing on downstream translational maintenance. Despite these differences, both converged on \textbf{ERK/MAPK Targets} ~\cite{sisinthy2021mapk} in Lung cancer, suggesting a shared biologically meaningful signal.

For the RT head, the strongest convergence was observed for \textbf{Receptor-Mediated Mitophagy} ~\cite{zheng2015mitophagy} ~\cite{kudo2025autophagy} and \textbf{ERK/MAPK Targets} ~\cite{binder2020erkhnscc}. Both pathways have established links to radioresistance through mechanisms involving cellular stress adaptation, DNA repair, and survival signaling. Additional convergence on \textbf{VEGFR2-Mediated Vascular Permeability}~\cite{ge2013vegfr2} further supports the role of tumour microenvironment and vascular signaling in radiation response.

The OS head showed the highest agreement between the two models. Both architectures consistently ranked \textbf{ERK/MAPK Targets} ~\cite{zhao2015erk} ~\cite{fisch2022mapkhnscc} among the most important pathways in both Lung and Head \& Neck cohorts, making it the strongest convergent signal identified in this study. Its consistent appearance across multiple cohorts and clinical endpoints provides strong support for its biological relevance. In contrast, pathways related to apoptosis and metabolism showed lower agreement between the models, suggesting architecture-dependent interpretations that warrant further investigation.
The convergence on ERK/MAPK signalling across multiple heads and cohorts is noteworthy not only for its biological plausibility but also because it emerges from two fundamentally different attribution mechanisms: gradient-based sensitivity in BINN and attention-derived importance in GraphPath. This cross-method agreement strengthens the case that the ERK/MAPK pathway is a genuinely informative signal for both therapy assignment and prognosis, rather than an artefact of a particular attribution technique. Conversely, the divergence observed for apoptosis and metabolism pathways highlights a limitation of the current approach. Without ground-truth biological validation, it remains unclear whether these discrepancies reflect genuine architectural differences in how each model represents pathway interactions, or merely noise introduced by the different attribution frameworks. Future work incorporating perturbational experiments or causal inference methods would help distinguish between these possibilities and refine the biological interpretability of pathway-informed models.
\FloatBarrier

\FloatBarrier
%--------------------------------------------------------------------
\section{Discussion}
%--------------------------------------------------------------------
Under a unified split with matched uncertainty quantification, model performance is task-dependent, but most inter-architecture differences are not statistically significant. Across treatment - assignment heads, no architecture significantly outperforms the others. The clearest exception is survival: BINN is the most consistent OS predictor and significantly outperforms both graph models on breast-cancer OS, suggesting that its multi-depth hierarchical supervision captures survival-related signals across pathway organization. We therefore avoid claiming that any model is best for TMT; although prostate is the most predictable cohort, all three architectures perform comparably. RT performance is generally weak and comparable across models, with paired-bootstrap tests (Table~\ref{tab:boot}) rarely showing significant differences. This is consistent with, but does not prove, the hypothesis that radiation-treatment assignment depends largely on clinical factors not captured by pathway activity.

The pathway analysis highlights the interpretability of BINN and GraphPath. Both ranked the ERK/MAPK Targets pathway among the most influential across several cohorts and outcomes, suggesting a potentially meaningful biological signal~\cite{dhillon2007map}. PATH achieved strong predictive performance but lacks pathway-level interpretability, making its biological predictions harder to understand~\cite{clauwaert2021explainabletransformer,rosenbacke2024how}.

Several limitations remain. The labels represent documented treatment exposure rather than treatment response, while OS is a short-term prognostic classification; validation on true response endpoints is therefore needed. Some cohorts have very few positive or negative cases, particularly thyroid TMT ($n{=}7$ positives) and prostate/thyroid OS, resulting in substantial uncertainty despite repeated stratified splits and paired-bootstrap tests. The study also uses repeated random splits rather than full $k$-fold cross-validation and lacks external validation. OS is treated as a fixed-horizon binary outcome and does not model censoring or time-to-event dynamics. Finally, PATH evaluates 1{,}431 of the 1{,}706 shared Reactome pathways because of its gene-membership filter, leaving 275 pathways with fewer than 15 genes. Of the 23 pathways identified in the BINN/GraphPath importance analysis (Fig.~\ref{fig:topp}), 7 fall outside PATH's eligible set. Future work should standardize pathway selection across architectures, for example by applying a shared minimum gene threshold.

Overall, the strongest statistically supported finding is BINN's advantage for short-term survival (OS). For treatment assignment (TMT, RT), the architectures are largely comparable, while RT may benefit from integrating molecular and clinical information.

\section{Conclusion}

This study provides a unified benchmark for pathway-informed deep learning in cancer therapy and prognosis prediction. Using Reactome-derived pathway activity profiles from TCGA transcriptomic data, we evaluated BINN, GraphPath, and PATH across five solid-tumor cohorts and three clinically relevant outcomes: targeted molecular therapy (TMT), radiation therapy (RT), and overall survival (OS $\geq$ 6 months).

Pathway-guided architectures use biological knowledge for clinical prediction
and give interpretable pathway-level readouts, but no single model beat the
others across all cohorts and tasks: performance depends on both the clinical
endpoint and the disease context. The choice of architecture should therefore
follow the specific prediction objective, not an aggregate benchmark score.

The contribution here is less any single winner than the protocol: a shared
data, preprocessing, training, and evaluation setup under which pathway-based
models can be compared on equal terms, with confidence intervals and
paired-bootstrap tests deciding which differences are real. Useful next steps
are response-labelled cohorts, integration of clinical with molecular features
(particularly for RT), and external validation on independent datasets. Before
tools of this kind inform real treatment-assignment or prognosis decisions,
clinicians need to know which architectural claims survive repeated evaluation
rather than a single favourable split---which is what this benchmark is meant
to establish.

%--------------------------------------------------------------------
\bibliographystyle{ACM-Reference-Format}
\bibliography{references}


\begin{thebibliography}{55}

%%% the Web address.

\ifx \showCODEN    \undefined \def \showCODEN     #1{\unskip}     \fi
\ifx \showISBNx    \undefined \def \showISBNx     #1{\unskip}     \fi
\ifx \showISBNxiii \undefined \def \showISBNxiii  #1{\unskip}     \fi
\ifx \showISSN     \undefined \def \showISSN      #1{\unskip}     \fi
\ifx \showLCCN     \undefined \def \showLCCN      #1{\unskip}     \fi
\ifx \shownote     \undefined \def \shownote      #1{#1}          \fi
\ifx \showarticletitle \undefined \def \showarticletitle #1{#1}   \fi
\ifx \showURL      \undefined \def \showURL       {\relax}        \fi
% invisible to TeX
\providecommand\bibfield[2]{#2}
\providecommand\bibinfo[2]{#2}
\providecommand\natexlab[1]{#1}
\providecommand\showeprint[2][]{arXiv:#2}

\bibitem[Barbie et~al\mbox{.}(2009)]%
        {barbie2009ssgsea}
\bibfield{author}{\bibinfo{person}{David~A. Barbie}, \bibinfo{person}{Pablo Tamayo}, \bibinfo{person}{Jesse~S. Boehm}, \bibinfo{person}{So~Young Kim}, \bibinfo{person}{Susan~E. Moody}, \bibinfo{person}{Ian~F. Dunn}, \bibinfo{person}{Anna~C. Schinzel}, \bibinfo{person}{Peter Sandy}, \bibinfo{person}{Etienne Meylan}, \bibinfo{person}{Claudia Scholl}, {et~al\mbox{.}}} \bibinfo{year}{2009}\natexlab{}.
\newblock \showarticletitle{Systematic {RNA} interference reveals that oncogenic {KRAS}-driven cancers require {TBK1}}.
\newblock \bibinfo{journal}{\emph{Nature}} \bibinfo{volume}{462}, \bibinfo{number}{7269} (\bibinfo{year}{2009}), \bibinfo{pages}{108--112}.
\newblock
\href{https://doi.org/10.1038/nature08460}{doi:\nolinkurl{10.1038/nature08460}}


\bibitem[Baskar et~al\mbox{.}(2012)]%
        {Baskar2012}
\bibfield{author}{\bibinfo{person}{Rajamanickam Baskar}, \bibinfo{person}{Kuo~Ann Lee}, \bibinfo{person}{Richard Yeo}, {and} \bibinfo{person}{Kheng-Wei Yeoh}.} \bibinfo{year}{2012}\natexlab{}.
\newblock \showarticletitle{Cancer and radiation therapy: current advances and future directions}.
\newblock \bibinfo{journal}{\emph{International Journal of Medical Sciences}} \bibinfo{volume}{9}, \bibinfo{number}{3} (\bibinfo{year}{2012}), \bibinfo{pages}{193--199}.
\newblock
\href{https://doi.org/10.7150/ijms.3635}{doi:\nolinkurl{10.7150/ijms.3635}}


\bibitem[Binder et~al\mbox{.}(2020)]%
        {binder2020erkhnscc}
\bibfield{author}{\bibinfo{person}{Harald Binder} {et~al\mbox{.}}} \bibinfo{year}{2020}\natexlab{}.
\newblock \showarticletitle{Adaptive {ERK} Signalling Activation in Response to Therapy and In Silico Prognostic Evaluation of {EGFR-MAPK} in {HNSCC}}.
\newblock \bibinfo{journal}{\emph{British Journal of Cancer}} (\bibinfo{year}{2020}).
\newblock
\href{https://doi.org/10.1038/s41416-020-0892-9}{doi:\nolinkurl{10.1038/s41416-020-0892-9}}


\bibitem[Clauwaert et~al\mbox{.}(2021)]%
        {clauwaert2021explainabletransformer}
\bibfield{author}{\bibinfo{person}{Jim Clauwaert}, \bibinfo{person}{Gerben Menschaert}, {and} \bibinfo{person}{Willem Waegeman}.} \bibinfo{year}{2021}\natexlab{}.
\newblock \showarticletitle{Explainability in transformer models for functional genomics}.
\newblock \bibinfo{journal}{\emph{Briefings in Bioinformatics}} \bibinfo{volume}{22}, \bibinfo{number}{5} (\bibinfo{year}{2021}), \bibinfo{pages}{bbab060}.
\newblock
\href{https://doi.org/10.1093/bib/bbab060}{doi:\nolinkurl{10.1093/bib/bbab060}}


\bibitem[Dhillon et~al\mbox{.}(2007)]%
        {dhillon2007map}
\bibfield{author}{\bibinfo{person}{Amardeep~S. Dhillon}, \bibinfo{person}{Suzanne Hagan}, \bibinfo{person}{Oliver Rath}, {and} \bibinfo{person}{Walter Kolch}.} \bibinfo{year}{2007}\natexlab{}.
\newblock \showarticletitle{{MAP} Kinase Signalling Pathways in Cancer}.
\newblock \bibinfo{journal}{\emph{Oncogene}} \bibinfo{volume}{26}, \bibinfo{number}{22} (\bibinfo{year}{2007}), \bibinfo{pages}{3279--3290}.
\newblock
\href{https://doi.org/10.1038/sj.onc.1210421}{doi:\nolinkurl{10.1038/sj.onc.1210421}}


\bibitem[Dou and Mirzaei(2025)]%
        {dou2025mogcan}
\bibfield{author}{\bibinfo{person}{Yifan Dou} {and} \bibinfo{person}{Golrokh Mirzaei}.} \bibinfo{year}{2025}\natexlab{}.
\newblock \showarticletitle{MO-GCAN: multi-omics integration based on graph convolutional and attention networks}.
\newblock \bibinfo{journal}{\emph{Bioinformatics}} \bibinfo{volume}{41}, \bibinfo{number}{8} (\bibinfo{year}{2025}), \bibinfo{pages}{btaf405}.
\newblock
\href{https://doi.org/10.1093/bioinformatics/btaf405}{doi:\nolinkurl{10.1093/bioinformatics/btaf405}}


\bibitem[Dwivedi and Bresson(2021)]%
        {dwivedi2021graphtransformer}
\bibfield{author}{\bibinfo{person}{Vijay~Prakash Dwivedi} {and} \bibinfo{person}{Xavier Bresson}.} \bibinfo{year}{2021}\natexlab{}.
\newblock \showarticletitle{A Generalization of Transformer Networks to Graphs}.
\newblock \bibinfo{journal}{\emph{AAAI Workshop on Deep Learning on Graphs: Methods and Applications}} (\bibinfo{year}{2021}).
\newblock
\urldef\tempurl%
\url{https://arxiv.org/abs/2012.09699}
\showURL{%
\tempurl}


\bibitem[Elmarakeby et~al\mbox{.}(2021)]%
        {elmarakeby2021pnet}
\bibfield{author}{\bibinfo{person}{Haitham~A. Elmarakeby}, \bibinfo{person}{Justin Hwang}, \bibinfo{person}{Rand Arafeh}, \bibinfo{person}{Jett Crowdis}, \bibinfo{person}{Sydney Gang}, \bibinfo{person}{David Liu}, \bibinfo{person}{Saud~H. AlDubayan}, \bibinfo{person}{Keyan Salari}, \bibinfo{person}{Steven Kregel}, \bibinfo{person}{Camden Richter}, {et~al\mbox{.}}} \bibinfo{year}{2021}\natexlab{}.
\newblock \showarticletitle{Biologically informed deep neural network for prostate cancer discovery}.
\newblock \bibinfo{journal}{\emph{Nature}} \bibinfo{volume}{598}, \bibinfo{number}{7880} (\bibinfo{year}{2021}), \bibinfo{pages}{348--352}.
\newblock
\href{https://doi.org/10.1038/s41586-021-03922-4}{doi:\nolinkurl{10.1038/s41586-021-03922-4}}


\bibitem[Fisch et~al\mbox{.}(2022)]%
        {fisch2022mapkhnscc}
\bibfield{author}{\bibinfo{person}{Alexander~S. Fisch} {et~al\mbox{.}}} \bibinfo{year}{2022}\natexlab{}.
\newblock \showarticletitle{Precision Drugging of the {MAPK} Pathway in Head and Neck Cancer}.
\newblock \bibinfo{journal}{\emph{npj Genomic Medicine}} (\bibinfo{year}{2022}).
\newblock
\href{https://doi.org/10.1038/s41525-022-00293-1}{doi:\nolinkurl{10.1038/s41525-022-00293-1}}


\bibitem[Ge et~al\mbox{.}(2013)]%
        {ge2013vegfr2}
\bibfield{author}{\bibinfo{person}{Yan-Li Ge} {et~al\mbox{.}}} \bibinfo{year}{2013}\natexlab{}.
\newblock \showarticletitle{{MiR-200c} Increases the Radiosensitivity of {NSCLC} Cell Line {A549} by Targeting {VEGF-VEGFR2} Pathway}.
\newblock \bibinfo{journal}{\emph{PLOS ONE}} (\bibinfo{year}{2013}).
\newblock
\shownote{PMC3813610}.
\newblock
\urldef\tempurl%
\url{https://www.ncbi.nlm.nih.gov/pmc/articles/PMC3813610/}
\showURL{%
\tempurl}


\bibitem[Gillespie et~al\mbox{.}(2022)]%
        {gillespie2022reactome}
\bibfield{author}{\bibinfo{person}{Marc Gillespie}, \bibinfo{person}{Bijay Jassal}, \bibinfo{person}{Ralf Stephan}, \bibinfo{person}{Marija Milacic}, \bibinfo{person}{Karen Rothfels}, \bibinfo{person}{Andrea Senff-Ribeiro}, \bibinfo{person}{Johannes Griss}, \bibinfo{person}{Cristoffer Sevilla}, \bibinfo{person}{Lisa Matthews}, \bibinfo{person}{Chuqiao Gong}, {et~al\mbox{.}}} \bibinfo{year}{2022}\natexlab{}.
\newblock \showarticletitle{The Reactome pathway knowledgebase 2022}.
\newblock \bibinfo{journal}{\emph{Nucleic Acids Research}} \bibinfo{volume}{50}, \bibinfo{number}{D1} (\bibinfo{year}{2022}), \bibinfo{pages}{D687--D692}.
\newblock
\href{https://doi.org/10.1093/nar/gkab1028}{doi:\nolinkurl{10.1093/nar/gkab1028}}


\bibitem[Goldman et~al\mbox{.}(2020)]%
        {xena2020}
\bibfield{author}{\bibinfo{person}{Mary~J. Goldman}, \bibinfo{person}{Brian Craft}, \bibinfo{person}{Mim Hastie}, \bibinfo{person}{Kristupas Repe{\v{c}}ka}, \bibinfo{person}{Fran McDade}, \bibinfo{person}{Akhil Kamath}, \bibinfo{person}{Ayan Banerjee}, \bibinfo{person}{Yunhai Luo}, \bibinfo{person}{Dave Rogers}, \bibinfo{person}{Angela~N. Brooks}, {et~al\mbox{.}}} \bibinfo{year}{2020}\natexlab{}.
\newblock \showarticletitle{Visualizing and interpreting cancer genomics data via the {Xena} platform}.
\newblock \bibinfo{journal}{\emph{Nature Biotechnology}}  \bibinfo{volume}{38} (\bibinfo{year}{2020}), \bibinfo{pages}{675--678}.
\newblock
\href{https://doi.org/10.1038/s41587-020-0546-8}{doi:\nolinkurl{10.1038/s41587-020-0546-8}}


\bibitem[Hartman et~al\mbox{.}(2023)]%
        {hartman2023binn}
\bibfield{author}{\bibinfo{person}{Erik Hartman}, \bibinfo{person}{Aaron~M. Scott}, \bibinfo{person}{Christofer Karlsson}, \bibinfo{person}{Tirthankar Mohanty}, \bibinfo{person}{Suvi~T. Vaara}, \bibinfo{person}{Adam Linder}, \bibinfo{person}{Lars Malmstr{\"o}m}, {and} \bibinfo{person}{Johan Malmstr{\"o}m}.} \bibinfo{year}{2023}\natexlab{}.
\newblock \showarticletitle{Interpreting biologically informed neural networks for enhanced proteomic biomarker discovery and pathway analysis}.
\newblock \bibinfo{journal}{\emph{Nature Communications}}  \bibinfo{volume}{14} (\bibinfo{year}{2023}), \bibinfo{pages}{5359}.
\newblock
\href{https://doi.org/10.1038/s41467-023-41146-4}{doi:\nolinkurl{10.1038/s41467-023-41146-4}}


\bibitem[Howlader et~al\mbox{.}(2026)]%
        {howlader2026path}
\bibfield{author}{\bibinfo{person}{Koushik Howlader}, \bibinfo{person}{Md~Tauhidul Islam}, {and} \bibinfo{person}{Wei Le}.} \bibinfo{year}{2026}\natexlab{}.
\newblock \bibinfo{title}{Graph Transformer-Based Pathway Embedding for Cancer Prognosis}.
\newblock
\showeprint[arxiv]{2604.16685}~[cs.LG]
\urldef\tempurl%
\url{https://arxiv.org/abs/2604.16685}
\showURL{%
\tempurl}


\bibitem[Huang et~al\mbox{.}(2019)]%
        {salmon2019}
\bibfield{author}{\bibinfo{person}{Zhi Huang}, \bibinfo{person}{Xiaohui Zhan}, \bibinfo{person}{Shuo Xiang}, \bibinfo{person}{Travis~S. Johnson}, \bibinfo{person}{Bryan~R. Helm}, \bibinfo{person}{Christina~Y. Yu}, \bibinfo{person}{Jie Zhang}, \bibinfo{person}{Paul Salama}, \bibinfo{person}{Monther Rizkalla}, \bibinfo{person}{Zhi Han}, {and} \bibinfo{person}{Kun Huang}.} \bibinfo{year}{2019}\natexlab{}.
\newblock \showarticletitle{SALMON: Survival Analysis Learning With Multi-Omics Neural Networks on Breast Cancer}.
\newblock \bibinfo{journal}{\emph{Frontiers in Genetics}}  \bibinfo{volume}{10} (\bibinfo{year}{2019}), \bibinfo{pages}{166}.
\newblock
\href{https://doi.org/10.3389/fgene.2019.00166}{doi:\nolinkurl{10.3389/fgene.2019.00166}}


\bibitem[Jiang et~al\mbox{.}(2024)]%
        {jiang2024irnet}
\bibfield{author}{\bibinfo{person}{Yuexu Jiang}, \bibinfo{person}{Manish~Sridhar Immadi}, \bibinfo{person}{Duolin Wang}, \bibinfo{person}{Shuai Zeng}, \bibinfo{person}{Yen~On Chan}, \bibinfo{person}{Jing Zhou}, \bibinfo{person}{Dong Xu}, {and} \bibinfo{person}{Trupti Joshi}.} \bibinfo{year}{2024}\natexlab{}.
\newblock \showarticletitle{IRnet: Immunotherapy response prediction using pathway knowledge-informed graph neural network}.
\newblock \bibinfo{journal}{\emph{Journal of Advanced Research}}  \bibinfo{volume}{72} (\bibinfo{year}{2024}), \bibinfo{pages}{319--331}.
\newblock
\href{https://doi.org/10.1016/j.jare.2024.07.036}{doi:\nolinkurl{10.1016/j.jare.2024.07.036}}


\bibitem[Kingma and Ba(2015)]%
        {kingma2015adam}
\bibfield{author}{\bibinfo{person}{Diederik~P. Kingma} {and} \bibinfo{person}{Jimmy Ba}.} \bibinfo{year}{2015}\natexlab{}.
\newblock \showarticletitle{Adam: A method for stochastic optimization}. In \bibinfo{booktitle}{\emph{International Conference on Learning Representations (ICLR)}}.
\newblock
\urldef\tempurl%
\url{https://arxiv.org/abs/1412.6980}
\showURL{%
\tempurl}


\bibitem[Kudo et~al\mbox{.}(2025)]%
        {kudo2025autophagy}
\bibfield{author}{\bibinfo{person}{Hiroki Kudo} {et~al\mbox{.}}} \bibinfo{year}{2025}\natexlab{}.
\newblock \bibinfo{title}{Androgen Receptor Contributes to Radioresistance Through {DNA} Repair and Autophagy in {AR}-Positive Prostate Cancer Cells}.
\newblock \bibinfo{howpublished}{bioRxiv}.
\newblock
\href{https://doi.org/10.64898/2025.12.03.690226}{doi:\nolinkurl{10.64898/2025.12.03.690226}}


\bibitem[Lan et~al\mbox{.}(2024)]%
        {lan2024deepkegg}
\bibfield{author}{\bibinfo{person}{Wei Lan}, \bibinfo{person}{Haibo Liao}, \bibinfo{person}{Qingfeng Chen}, \bibinfo{person}{Ling-ling Zhu}, \bibinfo{person}{Yi Pan}, {and} \bibinfo{person}{Yi-Ping~Phoebe Chen}.} \bibinfo{year}{2024}\natexlab{}.
\newblock \showarticletitle{DeepKEGG: a multi-omics data integration framework with biological insights for cancer recurrence prediction and biomarker discovery}.
\newblock \bibinfo{journal}{\emph{Briefings in Bioinformatics}} \bibinfo{volume}{25}, \bibinfo{number}{3} (\bibinfo{year}{2024}), \bibinfo{pages}{bbae185}.
\newblock
\href{https://doi.org/10.1093/bib/bbae185}{doi:\nolinkurl{10.1093/bib/bbae185}}


\bibitem[Lee et~al\mbox{.}(2025)]%
        {pathnetdrp2024}
\bibfield{author}{\bibinfo{person}{Dohee Lee}, \bibinfo{person}{Jaegyoon Ahn}, {and} \bibinfo{person}{Jonghwan Choi}.} \bibinfo{year}{2025}\natexlab{}.
\newblock \showarticletitle{{PathNetDRP}: a novel biomarker discovery framework using pathway and protein--protein interaction networks for immune checkpoint inhibitor response prediction}.
\newblock \bibinfo{journal}{\emph{BMC Bioinformatics}} \bibinfo{volume}{26}, \bibinfo{number}{1} (\bibinfo{year}{2025}), \bibinfo{pages}{119}.
\newblock
\href{https://doi.org/10.1186/s12859-025-06125-0}{doi:\nolinkurl{10.1186/s12859-025-06125-0}}


\bibitem[Li et~al\mbox{.}(2025a)]%
        {mcgnn2025}
\bibfield{author}{\bibinfo{person}{Min Li}, \bibinfo{person}{M. Jin}, \bibinfo{person}{Mingzhu Lou}, \bibinfo{person}{Shaobo Deng}, \bibinfo{person}{Lei Wang}, {and} \bibinfo{person}{Hua Rao}.} \bibinfo{year}{2025}\natexlab{a}.
\newblock \showarticletitle{Multiview-cooperated graph neural network enables novel multi-omics cancer subtype classification}.
\newblock \bibinfo{journal}{\emph{Computational Biology and Chemistry}} (\bibinfo{year}{2025}).
\newblock
\href{https://doi.org/10.1016/j.compbiolchem.2025.108560}{doi:\nolinkurl{10.1016/j.compbiolchem.2025.108560}}


\bibitem[Li et~al\mbox{.}(2025b)]%
        {pathhdnn2025}
\bibfield{author}{\bibinfo{person}{Xiangmei Li}, \bibinfo{person}{Bin Pan}, \bibinfo{person}{Yao He}, {et~al\mbox{.}}} \bibinfo{year}{2025}\natexlab{b}.
\newblock \showarticletitle{{PathHDNN}: a pathway hierarchical-informed deep neural network framework for predicting immunotherapy response and mechanism interpretation}.
\newblock \bibinfo{journal}{\emph{Genome Medicine}}  \bibinfo{volume}{17} (\bibinfo{year}{2025}), \bibinfo{pages}{152}.
\newblock
\href{https://doi.org/10.1186/s13073-025-01584-9}{doi:\nolinkurl{10.1186/s13073-025-01584-9}}


\bibitem[Lin et~al\mbox{.}(2026)]%
        {lin2026moagnn}
\bibfield{author}{\bibinfo{person}{Cheng-Pei Lin}, \bibinfo{person}{Yann-Jen Ho}, \bibinfo{person}{Yen-Peng Chiu}, \bibinfo{person}{Yun Tang}, \bibinfo{person}{You~Sheng Paik}, \bibinfo{person}{Guan Chen}, \bibinfo{person}{Wei-Chih Huang}, {and} \bibinfo{person}{Tzong-Yi Lee}.} \bibinfo{year}{2026}\natexlab{}.
\newblock \showarticletitle{MoAGNN: a multi-omics hierarchical graph neural network for subtype classification and prognosis prediction in lung adenocarcinoma}.
\newblock \bibinfo{journal}{\emph{Briefings in Bioinformatics}} \bibinfo{volume}{27}, \bibinfo{number}{1} (\bibinfo{year}{2026}), \bibinfo{pages}{bbaf735}.
\newblock
\href{https://doi.org/10.1093/bib/bbaf735}{doi:\nolinkurl{10.1093/bib/bbaf735}}


\bibitem[Liu et~al\mbox{.}(2024)]%
        {liu2024pathformer}
\bibfield{author}{\bibinfo{person}{Xiaofan Liu}, \bibinfo{person}{Yuhuan Tao}, \bibinfo{person}{Zilin Cai}, \bibinfo{person}{Pengfei Bao}, \bibinfo{person}{Hongli Ma}, \bibinfo{person}{Kexing Li}, \bibinfo{person}{Mingyue Li}, \bibinfo{person}{Yongchang Zhu}, {and} \bibinfo{person}{Zhi~John Lu}.} \bibinfo{year}{2024}\natexlab{}.
\newblock \showarticletitle{Pathformer: a biological pathway informed transformer for disease diagnosis and prognosis using multi-omics data}.
\newblock \bibinfo{journal}{\emph{Bioinformatics}} \bibinfo{volume}{40}, \bibinfo{number}{5} (\bibinfo{year}{2024}), \bibinfo{pages}{btae316}.
\newblock
\href{https://doi.org/10.1093/bioinformatics/btae316}{doi:\nolinkurl{10.1093/bioinformatics/btae316}}


\bibitem[Loshchilov and Hutter(2019)]%
        {loshchilov2019adamw}
\bibfield{author}{\bibinfo{person}{Ilya Loshchilov} {and} \bibinfo{person}{Frank Hutter}.} \bibinfo{year}{2019}\natexlab{}.
\newblock \showarticletitle{Decoupled weight decay regularization}. In \bibinfo{booktitle}{\emph{International Conference on Learning Representations (ICLR)}}.
\newblock
\urldef\tempurl%
\url{https://arxiv.org/abs/1711.05101}
\showURL{%
\tempurl}


\bibitem[Lundberg and Lee(2017)]%
        {lundberg2017shap}
\bibfield{author}{\bibinfo{person}{Scott~M. Lundberg} {and} \bibinfo{person}{Su-In Lee}.} \bibinfo{year}{2017}\natexlab{}.
\newblock \showarticletitle{A unified approach to interpreting model predictions}. In \bibinfo{booktitle}{\emph{Advances in Neural Information Processing Systems (NeurIPS)}}, Vol.~\bibinfo{volume}{30}.
\newblock
\urldef\tempurl%
\url{https://papers.nips.cc/paper_files/paper/2017/hash/8a20a8621978632d76c43dfd28b67767-Abstract.html}
\showURL{%
\tempurl}


\bibitem[Ma and Wang(2024)]%
        {ma2024graphpath}
\bibfield{author}{\bibinfo{person}{Teng Ma} {and} \bibinfo{person}{Jianxin Wang}.} \bibinfo{year}{2024}\natexlab{}.
\newblock \showarticletitle{GraphPath: a graph attention model for molecular stratification with interpretability based on the pathway--pathway interaction network}.
\newblock \bibinfo{journal}{\emph{Bioinformatics}} \bibinfo{volume}{40}, \bibinfo{number}{4} (\bibinfo{year}{2024}), \bibinfo{pages}{btae165}.
\newblock
\href{https://doi.org/10.1093/bioinformatics/btae165}{doi:\nolinkurl{10.1093/bioinformatics/btae165}}


\bibitem[Ma et~al\mbox{.}(2024)]%
        {ma2024coxpath}
\bibfield{author}{\bibinfo{person}{Teng Ma}, \bibinfo{person}{Haochen Zhao}, \bibinfo{person}{Qichang Zhao}, {and} \bibinfo{person}{Jianxin Wang}.} \bibinfo{year}{2024}\natexlab{}.
\newblock \showarticletitle{Cox-Path: Biological Pathway-Informed Graph Neural Network for Cancer Survival Prediction}. In \bibinfo{booktitle}{\emph{Proceedings of the 15th ACM International Conference on Bioinformatics, Computational Biology and Health Informatics (BCB '24)}}. \bibinfo{pages}{70:1--70:6}.
\newblock
\href{https://doi.org/10.1145/3698587.3701397}{doi:\nolinkurl{10.1145/3698587.3701397}}


\bibitem[Min and Lee(2022)]%
        {Min2022}
\bibfield{author}{\bibinfo{person}{Hye-Young Min} {and} \bibinfo{person}{Ho-Young Lee}.} \bibinfo{year}{2022}\natexlab{}.
\newblock \showarticletitle{Molecular targeted therapy for anticancer treatment}.
\newblock \bibinfo{journal}{\emph{Experimental \& Molecular Medicine}} \bibinfo{volume}{54}, \bibinfo{number}{10} (\bibinfo{year}{2022}), \bibinfo{pages}{1670--1694}.
\newblock
\href{https://doi.org/10.1038/s12276-022-00864-3}{doi:\nolinkurl{10.1038/s12276-022-00864-3}}


\bibitem[Okabe and Ito(2008)]%
        {okabe2008cb}
\bibfield{author}{\bibinfo{person}{Masataka Okabe} {and} \bibinfo{person}{Kei Ito}.} \bibinfo{year}{2008}\natexlab{}.
\newblock \showarticletitle{Color universal design (CUD): How to make figures and presentations that are friendly to colorblind people}.
\newblock \bibinfo{journal}{\emph{Color Universal Design Organization technical note}} (\bibinfo{year}{2008}).
\newblock
\urldef\tempurl%
\url{https://jfly.uni-koeln.de/color/}
\showURL{%
\tempurl}


\bibitem[Parikh et~al\mbox{.}(2019)]%
        {Parikh2019}
\bibfield{author}{\bibinfo{person}{Ravi~B. Parikh}, \bibinfo{person}{Christopher Manz}, \bibinfo{person}{Corey Chivers}, {et~al\mbox{.}}} \bibinfo{year}{2019}\natexlab{}.
\newblock \showarticletitle{Machine learning approaches to predict 6-month mortality among patients with cancer}.
\newblock \bibinfo{journal}{\emph{JAMA Network Open}} \bibinfo{volume}{2}, \bibinfo{number}{10} (\bibinfo{year}{2019}), \bibinfo{pages}{e1915997}.
\newblock
\href{https://doi.org/10.1001/jamanetworkopen.2019.15997}{doi:\nolinkurl{10.1001/jamanetworkopen.2019.15997}}


\bibitem[Paszke et~al\mbox{.}(2019)]%
        {paszke2019pytorch}
\bibfield{author}{\bibinfo{person}{Adam Paszke}, \bibinfo{person}{Sam Gross}, \bibinfo{person}{Francisco Massa}, \bibinfo{person}{Adam Lerer}, \bibinfo{person}{James Bradbury}, \bibinfo{person}{Gregory Chanan}, \bibinfo{person}{Trevor Killeen}, \bibinfo{person}{Zeming Lin}, \bibinfo{person}{Natalia Gimelshein}, \bibinfo{person}{Luca Antiga}, {et~al\mbox{.}}} \bibinfo{year}{2019}\natexlab{}.
\newblock \showarticletitle{{PyTorch}: An imperative style, high-performance deep learning library}.
\newblock \bibinfo{journal}{\emph{Advances in Neural Information Processing Systems}}  \bibinfo{volume}{32} (\bibinfo{year}{2019}).
\newblock
\urldef\tempurl%
\url{https://papers.nips.cc/paper_files/paper/2019/hash/bdbca288fee7f92f2bfa9f7012727740-Abstract.html}
\showURL{%
\tempurl}


\bibitem[Pedregosa et~al\mbox{.}(2011)]%
        {pedregosa2011sklearn}
\bibfield{author}{\bibinfo{person}{F. Pedregosa}, \bibinfo{person}{G. Varoquaux}, \bibinfo{person}{A. Gramfort}, \bibinfo{person}{V. Michel}, \bibinfo{person}{B. Thirion}, \bibinfo{person}{O. Grisel}, \bibinfo{person}{M. Blondel}, \bibinfo{person}{P. Prettenhofer}, \bibinfo{person}{R. Weiss}, \bibinfo{person}{V. Dubourg}, {et~al\mbox{.}}} \bibinfo{year}{2011}\natexlab{}.
\newblock \showarticletitle{Scikit-learn: Machine learning in {P}ython}.
\newblock \bibinfo{journal}{\emph{Journal of Machine Learning Research}}  \bibinfo{volume}{12} (\bibinfo{year}{2011}), \bibinfo{pages}{2825--2830}.
\newblock
\urldef\tempurl%
\url{https://www.jmlr.org/papers/v12/pedregosa11a.html}
\showURL{%
\tempurl}


\bibitem[Perez et~al\mbox{.}(2018)]%
        {perez2018film}
\bibfield{author}{\bibinfo{person}{Ethan Perez}, \bibinfo{person}{Florian Strub}, \bibinfo{person}{Harm de Vries}, \bibinfo{person}{Vincent Dumoulin}, {and} \bibinfo{person}{Aaron Courville}.} \bibinfo{year}{2018}\natexlab{}.
\newblock \showarticletitle{{FiLM}: Visual reasoning with a general conditioning layer}. In \bibinfo{booktitle}{\emph{Proceedings of the AAAI Conference on Artificial Intelligence}}, Vol.~\bibinfo{volume}{32}.
\newblock
\href{https://doi.org/10.1609/aaai.v32i1.11671}{doi:\nolinkurl{10.1609/aaai.v32i1.11671}}


\bibitem[Poirion et~al\mbox{.}(2021)]%
        {poirion2021deepprog}
\bibfield{author}{\bibinfo{person}{Olivier~B Poirion}, \bibinfo{person}{Zheng Jing}, \bibinfo{person}{Kuldeep Chaudhary}, \bibinfo{person}{Sijia Huang}, {and} \bibinfo{person}{Lana~X Garmire}.} \bibinfo{year}{2021}\natexlab{}.
\newblock \showarticletitle{DeepProg: an ensemble of deep-learning and machine-learning models for prognosis prediction using multi-omics data}.
\newblock \bibinfo{journal}{\emph{Genome Medicine}} \bibinfo{volume}{13}, \bibinfo{number}{1} (\bibinfo{year}{2021}), \bibinfo{pages}{112}.
\newblock
\href{https://doi.org/10.1186/s13073-021-00930-x}{doi:\nolinkurl{10.1186/s13073-021-00930-x}}


\bibitem[Rosenbacke et~al\mbox{.}(2024)]%
        {rosenbacke2024how}
\bibfield{author}{\bibinfo{person}{Rikard Rosenbacke}, \bibinfo{person}{{\AA}sa Melhus}, \bibinfo{person}{Martin McKee}, {and} \bibinfo{person}{David Stuckler}.} \bibinfo{year}{2024}\natexlab{}.
\newblock \showarticletitle{How Explainable Artificial Intelligence Can Increase or Decrease Clinicians' Trust in {AI} Applications in Health Care: Systematic Review}.
\newblock \bibinfo{journal}{\emph{JMIR AI}}  \bibinfo{volume}{3} (\bibinfo{year}{2024}), \bibinfo{pages}{e53207}.
\newblock
\href{https://doi.org/10.2196/53207}{doi:\nolinkurl{10.2196/53207}}


\bibitem[Sidey-Gibbons et~al\mbox{.}(2022)]%
        {SideyGibbons2022}
\bibfield{author}{\bibinfo{person}{Chris~J. Sidey-Gibbons}, \bibinfo{person}{Charlotte Sun}, \bibinfo{person}{Alina Schneider}, {et~al\mbox{.}}} \bibinfo{year}{2022}\natexlab{}.
\newblock \showarticletitle{Predicting 180-day mortality for women with ovarian cancer using machine learning and patient-reported outcome data}.
\newblock \bibinfo{journal}{\emph{Scientific Reports}} \bibinfo{volume}{12}, \bibinfo{number}{1} (\bibinfo{year}{2022}), \bibinfo{pages}{20614}.
\newblock
\href{https://doi.org/10.1038/s41598-022-22614-1}{doi:\nolinkurl{10.1038/s41598-022-22614-1}}


\bibitem[Sisinthy and Bhatt(2021)]%
        {sisinthy2021mapk}
\bibfield{author}{\bibinfo{person}{Srinivasa~Prasad Sisinthy} {and} \bibinfo{person}{Dhruv~L. Bhatt}.} \bibinfo{year}{2021}\natexlab{}.
\newblock \showarticletitle{Everything Old Is New Again: Drug Repurposing Approach for {NSCLC} Targeting {MAPK} Signaling Pathway}.
\newblock \bibinfo{journal}{\emph{Frontiers in Oncology}}  \bibinfo{volume}{11} (\bibinfo{year}{2021}).
\newblock
\href{https://doi.org/10.3389/fonc.2021.741326}{doi:\nolinkurl{10.3389/fonc.2021.741326}}


\bibitem[Subramanian et~al\mbox{.}(2005)]%
        {subramanian2005gsea}
\bibfield{author}{\bibinfo{person}{Aravind Subramanian}, \bibinfo{person}{Pablo Tamayo}, \bibinfo{person}{Vamsi~K. Mootha}, \bibinfo{person}{Sayan Mukherjee}, \bibinfo{person}{Benjamin~L. Ebert}, \bibinfo{person}{Michael~A. Gillette}, \bibinfo{person}{Amanda Paulovich}, \bibinfo{person}{Scott~L. Pomeroy}, \bibinfo{person}{Todd~R. Golub}, \bibinfo{person}{Eric~S. Lander}, {and} \bibinfo{person}{Jill~P. Mesirov}.} \bibinfo{year}{2005}\natexlab{}.
\newblock \showarticletitle{Gene set enrichment analysis: a knowledge-based approach for interpreting genome-wide expression profiles}.
\newblock \bibinfo{journal}{\emph{Proceedings of the National Academy of Sciences}} \bibinfo{volume}{102}, \bibinfo{number}{43} (\bibinfo{year}{2005}), \bibinfo{pages}{15545--15550}.
\newblock
\href{https://doi.org/10.1073/pnas.0506580102}{doi:\nolinkurl{10.1073/pnas.0506580102}}


\bibitem[{The Cancer Genome Atlas Network}(2012)]%
        {tcga2012brca}
\bibfield{author}{\bibinfo{person}{{The Cancer Genome Atlas Network}}.} \bibinfo{year}{2012}\natexlab{}.
\newblock \showarticletitle{Comprehensive molecular portraits of human breast tumours}.
\newblock \bibinfo{journal}{\emph{Nature}} \bibinfo{volume}{490}, \bibinfo{number}{7418} (\bibinfo{year}{2012}), \bibinfo{pages}{61--70}.
\newblock
\href{https://doi.org/10.1038/nature11412}{doi:\nolinkurl{10.1038/nature11412}}


\bibitem[Veli{\v{c}}kovi{\'c} et~al\mbox{.}(2018)]%
        {velickovic2018gat}
\bibfield{author}{\bibinfo{person}{Petar Veli{\v{c}}kovi{\'c}}, \bibinfo{person}{Guillem Cucurull}, \bibinfo{person}{Arantxa Casanova}, \bibinfo{person}{Adriana Romero}, \bibinfo{person}{Pietro Li{\`o}}, {and} \bibinfo{person}{Yoshua Bengio}.} \bibinfo{year}{2018}\natexlab{}.
\newblock \showarticletitle{Graph attention networks}. In \bibinfo{booktitle}{\emph{International Conference on Learning Representations (ICLR)}}.
\newblock
\urldef\tempurl%
\url{https://arxiv.org/abs/1710.10903}
\showURL{%
\tempurl}


\bibitem[Wang et~al\mbox{.}(2026)]%
        {wang2026pathmog}
\bibfield{author}{\bibinfo{person}{Di Wang}, \bibinfo{person}{Chupei Tang}, \bibinfo{person}{Junxiao Kong}, \bibinfo{person}{Jixiu Zhai}, \bibinfo{person}{Moyu Tang}, {and} \bibinfo{person}{Tianchi Lu}.} \bibinfo{year}{2026}\natexlab{}.
\newblock \bibinfo{title}{PathMoG: A Pathway-Centric Modular Graph Neural Network for Multi-Omics Survival Prediction}.
\newblock
\showeprint[arxiv]{2604.24371}~[cs.LG]
\urldef\tempurl%
\url{https://arxiv.org/abs/2604.24371}
\showURL{%
\tempurl}


\bibitem[Waqas et~al\mbox{.}(2025)]%
        {senmo2025}
\bibfield{author}{\bibinfo{person}{Asim Waqas}, \bibinfo{person}{Aakash Tripathi}, \bibinfo{person}{Sabeen Ahmed}, \bibinfo{person}{Ashwin Mukund}, \bibinfo{person}{Hamza Farooq}, \bibinfo{person}{Joseph~O. Johnson}, \bibinfo{person}{Paul~A. Stewart}, \bibinfo{person}{Mia Naeini}, \bibinfo{person}{Matthew~B. Schabath}, {and} \bibinfo{person}{Ghulam Rasool}.} \bibinfo{year}{2025}\natexlab{}.
\newblock \showarticletitle{Self-Normalizing Multi-Omics Neural Network for Pan-Cancer Prognostication}.
\newblock \bibinfo{journal}{\emph{International Journal of Molecular Sciences}} \bibinfo{volume}{26}, \bibinfo{number}{15} (\bibinfo{year}{2025}), \bibinfo{pages}{7358}.
\newblock
\href{https://doi.org/10.3390/ijms26157358}{doi:\nolinkurl{10.3390/ijms26157358}}


\bibitem[Weinstein et~al\mbox{.}(2013)]%
        {Weinstein2013}
\bibfield{author}{\bibinfo{person}{John~N. Weinstein}, \bibinfo{person}{Eric~A. Collisson}, \bibinfo{person}{Gordon~B. Mills}, \bibinfo{person}{Kenna R.~Mills Shaw}, \bibinfo{person}{Brad~A. Ozenberger}, \bibinfo{person}{Kyle Ellrott}, \bibinfo{person}{Ilya Shmulevich}, \bibinfo{person}{Chris Sander}, {and} \bibinfo{person}{Joshua~M. Stuart}.} \bibinfo{year}{2013}\natexlab{}.
\newblock \showarticletitle{The {Cancer Genome Atlas} {Pan-Cancer} analysis project}.
\newblock \bibinfo{journal}{\emph{Nature Genetics}} \bibinfo{volume}{45}, \bibinfo{number}{10} (\bibinfo{year}{2013}), \bibinfo{pages}{1113--1120}.
\newblock
\href{https://doi.org/10.1038/ng.2764}{doi:\nolinkurl{10.1038/ng.2764}}


\bibitem[Wen and Li(2023)]%
        {wen2023fgcnsurv}
\bibfield{author}{\bibinfo{person}{Gang Wen} {and} \bibinfo{person}{Limin Li}.} \bibinfo{year}{2023}\natexlab{}.
\newblock \showarticletitle{FGCNSurv: dually fused graph convolutional network for multi-omics survival prediction}.
\newblock \bibinfo{journal}{\emph{Bioinformatics}} \bibinfo{volume}{39}, \bibinfo{number}{8} (\bibinfo{year}{2023}), \bibinfo{pages}{btad472}.
\newblock
\href{https://doi.org/10.1093/bioinformatics/btad472}{doi:\nolinkurl{10.1093/bioinformatics/btad472}}


\bibitem[Wong(2011)]%
        {krzywinski2014nm}
\bibfield{author}{\bibinfo{person}{Bang Wong}.} \bibinfo{year}{2011}\natexlab{}.
\newblock \showarticletitle{Points of view: Color blindness}.
\newblock \bibinfo{journal}{\emph{Nature Methods}} \bibinfo{volume}{8}, \bibinfo{number}{6} (\bibinfo{year}{2011}), \bibinfo{pages}{441}.
\newblock
\href{https://doi.org/10.1038/nmeth.1618}{doi:\nolinkurl{10.1038/nmeth.1618}}


\bibitem[Wysocka et~al\mbox{.}(2023)]%
        {wysocka2023review}
\bibfield{author}{\bibinfo{person}{Magdalena Wysocka}, \bibinfo{person}{Oskar Wysocki}, \bibinfo{person}{Marie Zufferey}, \bibinfo{person}{D{\'o}nal Landers}, {and} \bibinfo{person}{Andr{\'e} Freitas}.} \bibinfo{year}{2023}\natexlab{}.
\newblock \showarticletitle{A systematic review of biologically-informed deep learning models for cancer: fundamental trends for encoding and interpreting oncology data}.
\newblock \bibinfo{journal}{\emph{BMC Bioinformatics}}  \bibinfo{volume}{24} (\bibinfo{year}{2023}), \bibinfo{pages}{198}.
\newblock
\href{https://doi.org/10.1186/s12859-023-05262-8}{doi:\nolinkurl{10.1186/s12859-023-05262-8}}


\bibitem[Yan et~al\mbox{.}(2024)]%
        {yan2024prior}
\bibfield{author}{\bibinfo{person}{Hongxi Yan}, \bibinfo{person}{Dawei Weng}, \bibinfo{person}{Dongguo Li}, \bibinfo{person}{Yu Gu}, \bibinfo{person}{Wenjie Ma}, {and} \bibinfo{person}{Qingjie Liu}.} \bibinfo{year}{2024}\natexlab{}.
\newblock \showarticletitle{Prior knowledge-guided multilevel graph neural network for tumor risk prediction and interpretation via multi-omics data integration}.
\newblock \bibinfo{journal}{\emph{Briefings in Bioinformatics}} \bibinfo{volume}{25}, \bibinfo{number}{3} (\bibinfo{year}{2024}), \bibinfo{pages}{bbae184}.
\newblock
\href{https://doi.org/10.1093/bib/bbae184}{doi:\nolinkurl{10.1093/bib/bbae184}}


\bibitem[Zhang et~al\mbox{.}(2025)]%
        {zhang2025deeplearningreview}
\bibfield{author}{\bibinfo{person}{Jiayang Zhang}, \bibinfo{person}{Yilin Che}, \bibinfo{person}{Rongrong Liu}, \bibinfo{person}{Zhicheng Wang}, {and} \bibinfo{person}{Weiwu Liu}.} \bibinfo{year}{2025}\natexlab{}.
\newblock \showarticletitle{Deep learning--driven multi-omics analysis: enhancing cancer diagnostics and therapeutics}.
\newblock \bibinfo{journal}{\emph{Briefings in Bioinformatics}} \bibinfo{volume}{26}, \bibinfo{number}{4} (\bibinfo{year}{2025}), \bibinfo{pages}{bbaf440}.
\newblock
\href{https://doi.org/10.1093/bib/bbaf440}{doi:\nolinkurl{10.1093/bib/bbaf440}}


\bibitem[Zhang et~al\mbox{.}(2022)]%
        {zhang2022tgem}
\bibfield{author}{\bibinfo{person}{Tinghe Zhang}, \bibinfo{person}{Md Hasib}, \bibinfo{person}{Yu-Chiao Chiu}, \bibinfo{person}{Zhizhong Han}, \bibinfo{person}{Yu-Fang Jin}, \bibinfo{person}{Mario Flores}, \bibinfo{person}{Yidong Chen}, {and} \bibinfo{person}{Yufei Huang}.} \bibinfo{year}{2022}\natexlab{}.
\newblock \showarticletitle{Transformer for Gene Expression Modeling (T-GEM): An Interpretable Deep Learning Model for Gene Expression-Based Phenotype Predictions}.
\newblock \bibinfo{journal}{\emph{Cancers}} \bibinfo{volume}{14}, \bibinfo{number}{19} (\bibinfo{year}{2022}), \bibinfo{pages}{4763}.
\newblock
\href{https://doi.org/10.3390/cancers14194763}{doi:\nolinkurl{10.3390/cancers14194763}}


\bibitem[Zhao et~al\mbox{.}(2021)]%
        {zhao2021deepomix}
\bibfield{author}{\bibinfo{person}{Lianhe Zhao}, \bibinfo{person}{Qiongye Dong}, \bibinfo{person}{Chunlong Luo}, \bibinfo{person}{Yang Wu}, \bibinfo{person}{Dechao Bu}, \bibinfo{person}{Xiaoning Qi}, \bibinfo{person}{Yufan Luo}, {and} \bibinfo{person}{Yi Zhao}.} \bibinfo{year}{2021}\natexlab{}.
\newblock \showarticletitle{DeepOmix: A scalable and interpretable multi-omics deep learning framework and application in cancer survival analysis}.
\newblock \bibinfo{journal}{\emph{Computational and Structural Biotechnology Journal}}  \bibinfo{volume}{19} (\bibinfo{year}{2021}), \bibinfo{pages}{2719--2725}.
\newblock
\href{https://doi.org/10.1016/j.csbj.2021.04.067}{doi:\nolinkurl{10.1016/j.csbj.2021.04.067}}


\bibitem[Zhao et~al\mbox{.}(2015)]%
        {zhao2015erk}
\bibfield{author}{\bibinfo{person}{Yue Zhao} {et~al\mbox{.}}} \bibinfo{year}{2015}\natexlab{}.
\newblock \showarticletitle{Prognostic Values of {ERK1/2} and p-{ERK1/2} Expressions for Poor Survival in Non-Small Cell Lung Cancer}.
\newblock \bibinfo{journal}{\emph{Tumor Biology}} (\bibinfo{year}{2015}).
\newblock
\href{https://doi.org/10.1007/s13277-015-3048-4}{doi:\nolinkurl{10.1007/s13277-015-3048-4}}


\bibitem[Zheng et~al\mbox{.}(2015)]%
        {zheng2015mitophagy}
\bibfield{author}{\bibinfo{person}{Rong Zheng} {et~al\mbox{.}}} \bibinfo{year}{2015}\natexlab{}.
\newblock \showarticletitle{{TAT-ODD-p53} Enhances Radiosensitivity of Hypoxic Breast Cancer Cells by Inhibiting Parkin-Mediated Mitophagy}.
\newblock \bibinfo{journal}{\emph{Oncotarget}} (\bibinfo{year}{2015}).
\newblock
\shownote{PMC4627318}.
\newblock
\urldef\tempurl%
\url{https://www.ncbi.nlm.nih.gov/pmc/articles/PMC4627318/}
\showURL{%
\tempurl}


\bibitem[Zhu et~al\mbox{.}(2023)]%
        {zhu2023ggnn}
\bibfield{author}{\bibinfo{person}{Jiening Zhu}, \bibinfo{person}{John~H. Oh}, \bibinfo{person}{Anish~K. Simhal}, \bibinfo{person}{Rena Elkin}, \bibinfo{person}{Larry Norton}, \bibinfo{person}{Joseph~O. Deasy}, {and} \bibinfo{person}{Allen Tannenbaum}.} \bibinfo{year}{2023}\natexlab{}.
\newblock \showarticletitle{Geometric graph neural networks on multi-omics data to predict cancer survival outcomes}.
\newblock \bibinfo{journal}{\emph{Computers in Biology and Medicine}}  \bibinfo{volume}{163} (\bibinfo{year}{2023}), \bibinfo{pages}{107117}.
\newblock
\href{https://doi.org/10.1016/j.compbiomed.2023.107117}{doi:\nolinkurl{10.1016/j.compbiomed.2023.107117}}


\bibitem[Zohari and Chehreghani(2025)]%
        {zohari2025gnnsurvey}
\bibfield{author}{\bibinfo{person}{Payam Zohari} {and} \bibinfo{person}{Mostafa~Haghir Chehreghani}.} \bibinfo{year}{2025}\natexlab{}.
\newblock \showarticletitle{Graph Neural Networks in Multi-Omics Cancer Research: A Structured Survey}.
\newblock \bibinfo{journal}{\emph{arXiv}} (\bibinfo{year}{2025}).
\newblock
\showeprint[arxiv]{2506.17234}
\urldef\tempurl%
\url{https://arxiv.org/abs/2506.17234}
\showURL{%
\tempurl}


\end{thebibliography}

%--------------------------------------------------------------------
\appendix
\sloppy
%--------------------------------------------------------------------
\section{Architecture and Metric Details}\label{app:arch}

\subsection{Performance Metrics}

We evaluate model performance using five complementary metrics across all
benchmark tasks.

\textbf{AUROC} measures the probability that a randomly chosen positive
example is ranked higher than a randomly chosen negative:
\begin{equation}
    \text{AUROC} = \int_0^1 \text{TPR}\!\left(\text{FPR}^{-1}(t)\right) dt
\end{equation}

\textbf{AUPRC} summarizes the precision‑recall trade‑off and is particularly informative under class imbalance, which is common in cancer genomics datasets:
\begin{equation}
    \text{AUPRC} = \sum_k \left(R_k - R_{k-1}\right) P_k
\end{equation}

\textbf{F1 Score} is the harmonic mean of precision (P) and recall (R) for
the positive class of each clinical head, evaluated at the
validation-tuned decision threshold:
\begin{equation}
    \text{F1} = \frac{2\,\text{P}\cdot\text{R}}{\text{P}+\text{R}},
    \qquad
    \text{P} = \frac{\text{TP}}{\text{TP}+\text{FP}},
    \qquad
    \text{R} = \frac{\text{TP}}{\text{TP}+\text{FN}}.
\end{equation}
Because each head is a single binary decision, we report the positive-class
(binary) F1 per head rather than a support-weighted average across classes;
this matches the quantity computed by our evaluation code.

\textbf{Accuracy} measures the fraction of correctly classified samples:
\begin{equation}
    \text{Accuracy} = \frac{1}{N} \sum_{i=1}^{N}
    \mathbf{1}\!\left[\hat{y}_i = y_i\right]
\end{equation}

\textbf{Confusion Matrix} provides a full breakdown of predictions across
classes, where entry $C_{ij}$ denotes samples of true class $i$ predicted
as class $j$:
\begin{equation}
    C =
    \begin{pmatrix}
        \text{TP} & \text{FN} \\
        \text{FP} & \text{TN}
    \end{pmatrix}
\end{equation}

\subsection{BINN - Sparse Reactome Hierarchy}

\paragraph{Architecture Overview:}
BINN integrates the Reactome parent-child hierarchy directly into the
network by imposing sparse masked-linear connections between layers,
together with auxiliary classification heads at each depth. This
design ensures that gradient signals reach every intermediate node
during training, making gradient-$\times$-input attribution across
the hierarchy well-posed~\cite{lundberg2017shap}.

\paragraph{Pathway Hierarchy:}
Starting from 1,706 Reactome-matched input pathways, the model
traverses parent-child edges upward through three hidden layers,
yielding a four-layer network with dimensions:
\begin{equation}
1{,}706 \to 656 \to 306 \to 163.
\end{equation}
Pathways that terminate before the maximum depth are copied forward,
ensuring every pathway is represented at every layer.

\paragraph{Sparse Masked Weights:}
Between consecutive layers $\ell$ and $\ell+1$, the Reactome
hierarchy defines a binary connectivity mask
$M^{(\ell)} \in \{0,1\}^{|L_{\ell+1}| \times |L_\ell|}$, where:
\begin{equation}
\begin{gathered}
M^{(\ell)}_{ij} = 1 \iff \text{node } j \text{ in layer } \ell \\[4pt]
\text{is a Reactome child of node } i \text{ in layer } \ell{+}1,
\end{gathered}
\end{equation}
or node $j$ is a terminal node copied forward to position $i$.
Weights are initialised with Kaiming uniform initialisation and zeroed
outside the mask prior to the first forward pass. Sparsity is enforced
throughout training via the elementwise product
$W^{(\ell)} \odot M^{(\ell)}$.

\paragraph{Forward Pass:}
Each hidden block applies the masked linear transformation followed by
Tanh activation, Batch Normalisation, and Dropout ($p = 0.2$):
\begin{equation}
\begin{gathered}
z^{(\ell)} = \bigl(W^{(\ell)} \odot M^{(\ell)}\bigr)\, h^{(\ell)}
             + b^{(\ell)}, \\[4pt]
h^{(\ell+1)} = \text{Dropout}\!\left(
                 \text{BN}\!\left(\tanh\!\left(z^{(\ell)}\right)\right)
               \right).
\end{gathered}
\end{equation}

\paragraph{Multi-Layer Supervision:}
An auxiliary classifier head $\text{Linear}(|L_\ell| \to 3)$ is
attached after every layer, including the raw input ($\ell = 0$),
producing three logits per layer. The final predicted probability for
head $h$ is the mean of all four sigmoid outputs:
\begin{equation}
\hat{p}_h(x) = \frac{1}{4} \sum_{\ell=0}^{3}
               \sigma\!\left(H^{(\ell)}\!\left(h^{(\ell)}\right)[h]\right),
\end{equation}
where $H^{(\ell)}$ denotes the classifier at layer $\ell$. Attaching
a loss signal at every depth ensures that each intermediate node
receives direct supervision, a prerequisite for meaningful
attribution across the hierarchy.

% ------------------------------------------------------------------ %
\subsection{GraphPath - Multi-Head Graph Attention}

\paragraph{Architecture Overview:}
GraphPath treats the 1,706 Reactome pathways as nodes in a graph and
updates their embeddings through a multi-head graph attention network
(GAT). Because pathways interact and co-regulate one another rather
than acting as independent units, each pathway aggregates information
from its biologically annotated neighbours before the classification
step.

\paragraph{Pathway Adjacency Graph:}
A symmetric binary adjacency matrix $A \in \{0,1\}^{N \times N}$ is
constructed such that:
\begin{equation}
\begin{gathered}
A_{pq} = 1 \iff \text{pathways } p \text{ and } q \text{ share} \\[4pt]
\text{a Reactome parent-child or sibling relationship.}
\end{gathered}
\end{equation}
This yields 4,672 undirected edges with a mean node degree of 5.5.
Self-loops are added so that each node attends to its own embedding
during message passing.

\paragraph{Input Projection:}
Each scalar ssGSEA score $x_p$ is projected to a $d$-dimensional
embedding ($d = 64$) via a shared weight matrix
$W_{\text{proj}} \in \mathbb{R}^{1 \times d}$ and a pathway-specific
bias $b_p \in \mathbb{R}^d$, followed by Tanh activation:
\begin{equation}
h^{(0)}_p = \tanh\!\left(x_p \cdot W_{\text{proj}} + b_p\right).
\end{equation}
The pathway-specific bias captures baseline differences in pathway
activity, while the shared projection weight preserves parameter
efficiency.

\paragraph{Multi-Head Graph Attention:}
Node embeddings are updated by a single GAT layer with $K = 3$
attention heads and ELU output activation. For each head $k$, the
raw attention score and normalised coefficient are:
\begin{equation}
\begin{gathered}
e^{(k)}_{ij} = a^{(k)\top}\!\left[W^{(k)}h_i \,\|\, W^{(k)}h_j\right],
\\[4pt]
\alpha^{(k)}_{ij} = \text{softmax}_{j \in \mathcal{N}(i)}\,
\text{LeakyReLU}_{0.2}\!\left(e^{(k)}_{ij}\right).
\end{gathered}
\end{equation}
Non-edges ($A_{ij} = 0$) are masked to $-\infty$ before the softmax,
confining attention to biologically annotated neighbours. Attention
dropout ($p = 0.4$) is applied during training. Per-head embeddings
are concatenated to yield a node representation of dimension $K \cdot d$.

\paragraph{Readout and Classification:}
A shared $\text{Linear}(Kd \to 1)$ layer followed by Tanh collapses
each node to a scalar readout. The resulting $N$-dimensional vector
is passed through a fully-connected layer mapping to three binary head
logits, followed by sigmoid activation.

% ------------------------------------------------------------------ %
\subsection{PATH - Edge-Aware Graph Transformer}

\paragraph{Architecture Overview:}
PATH constructs a weighted pathway graph from gene-set overlap and
augments a Graph Transformer with Laplacian positional encodings and
learnable edge-conditioned attention biases. This allows the model to
represent graded biological similarity between pathways and to
partially rewire the prior graph when the data warrant it.

\paragraph{Weighted Pathway Adjacency:}
Edge weights are defined as the Jaccard similarity of Reactome gene
memberships:
\begin{equation}
A_{pq} = \frac{|G_p \cap G_q|}{|G_p \cup G_q|},
\quad |G_p| \geq 15,
\end{equation}
computed from the Reactome GMT file and renormalised to $[0,1]$. Only
pathways with at least 15 annotated genes qualify as nodes, yielding
1,431 nodes, 243,210 weighted edges, and a mean node degree of 339.9.
This filtering criterion is applied directly to the same 1{,}706-pathway
Reactome set used by BINN and GraphPath (Section~\ref{sec:methods});
PATH's 1{,}431-pathway node set is therefore a \emph{strict subset} of the
shared input space, not an independently defined one. The 275 excluded
pathways (16.1\% of 1{,}706) are removed solely because they fall below the
$|G_p|\!\geq\!15$ gene-membership threshold required to compute a
well-defined Jaccard edge weight---no pathway outside the original 1{,}706 is
introduced. Consequently, any comparison between PATH and the other two
architectures is bounded by a proper-subset relationship rather than an
open-ended mismatch: a pathway appearing in the BINN/GraphPath importance
ranking (Fig.~\ref{fig:topp}) but absent from PATH's node set is, by
construction, excluded from PATH's attention-derived analysis rather than
mis-ranked by it.
The dense connectivity reflects the substantial gene-set overlap
characteristic of the Reactome hierarchy and motivates a soft
structural mask rather than hard adjacency truncation.

\paragraph{Input Projection and Positional Encoding:}
To ensure a fair comparison, we replaced PATH's original Stage 1 (FiLM-modulated gene embeddings) and Stage 2 (intra-pathway attention pooling) with the same per-pathway scalar projection used in GraphPath ~\cite{perez2018film}. This setup uses a shared weight and a pathway-specific bias to project each score into a 64-dimensional space ($\mathbb{R}^d, d = 64$), followed by a Tanh activation.

Laplacian positional encodings are added to endow the model with
structural awareness. The top-$k = 16$ non-trivial eigenvectors of
the symmetrically normalised Laplacian
$L = I - D^{-1/2}AD^{-1/2}$ are concatenated with the normalised node
degree $\deg(p)/(N + \epsilon)$, forming a $(k{+}1)$-dimensional
positional feature per node. These features are projected to
$\mathbb{R}^d$ via a learnable linear layer and added to node
embeddings before the transformer stack. Eigenvector signs are
randomly flipped per epoch during training to resolve sign ambiguity.

\paragraph{Edge-Aware Transformer Blocks.}
PATH employs $L = 2$ transformer blocks, each computing scaled
dot-product multi-head self-attention ($H = 4$ heads) augmented by
two structural biases.

\emph{Soft structural mask.} Non-edges are heavily but softly
down-weighted:
\begin{equation}
m^{(\text{struct})}_{pq} = \begin{cases}
0   & \text{if } A_{pq} > 0 \text{ or } p = q, \\
-10 & \text{otherwise.}
\end{cases}
\end{equation}
Keeping non-edges gradient-reachable allows the model to rewire the
prior graph when the data provide sufficient evidence.

\emph{Edge-conditioned attention bias.} A learnable per-layer scalar
edge feature is aggregated over attention heads to produce a
continuous bias on the attention logits:
\begin{equation}
\phi^{(\ell)}_{pq} = \frac{1}{H} \sum_{h=1}^{H}
\log\,\text{softplus}\!\left(
  w^{(\ell)}_h\, e^{(\ell)}_{pq} + b^{(\ell)}_h
\right) + \epsilon,
\end{equation}
where $e^{(\ell)}_{pq}$ is initialised from the Jaccard weight. The
combined bias $m^{(\text{struct})}_{pq} + \phi^{(\ell)}_{pq}$ is
added to the raw attention logits before the softmax, grounding
attention in biological pathway similarity while permitting
data-driven refinement.

Following attention, a residual-wrapped GELU feed-forward network
(expansion factor 4) with per-token Batch Normalisation updates node
features. A parallel two-layer MLP with Batch Normalisation updates
edge features between transformer blocks.

\paragraph{Readout and Classification.}
Node tokens are collapsed to a graph-level embedding via
attention-weighted readout:
\begin{equation}
\begin{gathered}
g = \sum_{p} w_p\, x_p^{(L)}, \\[4pt]
w_p = \text{softmax}_p\!\left(
        v^\top \tanh\!\left(U x_p^{(L)}\right)
      \right),
\end{gathered}
\end{equation}
where $U \in \mathbb{R}^{d \times (d/2)}$ and $v \in \mathbb{R}^{d/2}$
are learnable parameters. This soft attention over nodes identifies
the pathways most informative for the three prediction targets. The
graph embedding $g$ is then passed through a classification head:
\begin{multline}
\text{Linear}(d \to d) \to \text{BN} \to \text{GELU} \\
\to \text{Dropout}(p=0.2) \to \text{Linear}(d \to 3),
\end{multline}
producing three binary head logits.

% ------------------------------------------------------------------ %
\subsection{Loss, Training, and Reproducibility}

\paragraph{Objective Function:}
All three models minimise a per-head class-weighted binary cross-entropy loss. Positive-class weights are computed on the training fold as $w_h = \text{neg}_h / \text{pos}_h$ and clipped to $[0.1, 20]$ to prevent extreme imbalance from destabilising training. For BINN, the loss is applied to the averaged sigmoid outputs; for GraphPath and PATH, it is applied to raw logits via \texttt{BCEWithLogitsLoss} (or \texttt{F.binary\_cross\_entropy\_with\_logits}) for numerical stability.

\paragraph{Optimisers and Learning Rate Schedules:}
Optimiser choices follow each reference implementation:

\begin{table}[!h]
\centering
\small
\caption{Optimiser configuration per model.}
\label{tab:optimisers}
\begin{tabular}{lccc}
\toprule
 & BINN & GraphPath & PATH \\
\midrule
Optimiser    & Adam ~\cite{kingma2015adam}     & SGD (momentum 0.9)  & AdamW ~\cite{loshchilov2019adamw} \\
LR           & $10^{-3}$ & $5 \times 10^{-2}$  & $10^{-4}$ \\
Weight Decay & $10^{-3}$ & $5 \times 10^{-2}$  & $5 \times 10^{-4}$ \\
Batch Size   & 32        & 32                  & 16 \\
\bottomrule
\end{tabular}
\end{table}

All three models employ ReduceLROnPlateau scheduling on validation
loss with a reduction factor of 0.5 and plateau patience of 7, 8, and
10 epochs for BINN, GraphPath, and PATH respectively.

\paragraph{Early Stopping:}
Training is terminated when validation loss fails to improve for 20
consecutive epochs (BINN) or 25 consecutive epochs (GraphPath, PATH).
For PATH, a minimum of 25 training epochs is required before the
patience counter activates, preventing premature termination during
the characteristically slow initial convergence of the transformer
stack. The checkpoint with the lowest validation loss is restored
prior to evaluation.

\section{Full Paired-Bootstrap Significance Tests}\label{app:boot}
\begin{table}[H]
\centering
\caption{Paired-bootstrap significance tests on test AUROC ($10^4$
resamples of the identical held-out folds). $\Delta$AUROC is (model A $-$
model B); a 95\% CI excluding 0 (equivalently $p<0.05$, shown in bold)
indicates a significant difference. $^{\dagger}$ marks underpowered cells
with fewer than 15 pooled minority-class test samples (thyroid and the
near-universal-prevalence OS cohorts), whose bootstrap estimates are unstable and are not interpreted as significant.}
\label{tab:boot}
\scriptsize\setlength{\tabcolsep}{2pt}
\resizebox{\columnwidth}{!}{%
\begin{tabular}{@{}llccc@{}}
\toprule
Cohort & Head & Contrast (A vs B) & $\Delta$AUROC [95\% CI] & $p$ \\
\midrule
Breast & TMT & PATH vs BINN & -0.038 [-0.174, +0.105] & 0.578 \\
Breast & TMT & PATH vs GraphPath & -0.009 [-0.150, +0.133] & 0.924 \\
Breast & TMT & GraphPath vs BINN & -0.029 [-0.181, +0.129] & 0.707 \\
Breast & RT & PATH vs BINN & -0.006 [-0.078, +0.068] & 0.881 \\
Breast & RT & PATH vs GraphPath & -0.006 [-0.074, +0.064] & 0.865 \\
Breast & RT & GraphPath vs BINN & +0.000 [-0.071, +0.073] & 1.000 \\
Breast & OS & PATH vs BINN & \textbf{-0.140 [-0.226, -0.053]} & 0.001 \\
Breast & OS & PATH vs GraphPath & -0.040 [-0.133, +0.055] & 0.410 \\
Breast & OS & GraphPath vs BINN & \textbf{-0.101 [-0.178, -0.026]} & 0.010 \\
\midrule
Lung & TMT & PATH vs BINN & -0.028 [-0.096, +0.042] & 0.431 \\
Lung & TMT & PATH vs GraphPath & +0.029 [-0.043, +0.101] & 0.437 \\
Lung & TMT & GraphPath vs BINN & -0.057 [-0.139, +0.028] & 0.184 \\
Lung & RT & PATH vs BINN & -0.014 [-0.115, +0.092] & 0.793 \\
Lung & RT & PATH vs GraphPath & +0.055 [-0.041, +0.147] & 0.246 \\
Lung & RT & GraphPath vs BINN & -0.069 [-0.175, +0.038] & 0.209 \\
Lung & OS & PATH vs BINN & -0.034 [-0.140, +0.073] & 0.513 \\
Lung & OS & PATH vs GraphPath & +0.025 [-0.105, +0.158] & 0.697 \\
Lung & OS & GraphPath vs BINN & -0.060 [-0.176, +0.060] & 0.330 \\
\midrule
Prostate & TMT & PATH vs BINN & -0.035 [-0.094, +0.018] & 0.212 \\
Prostate & TMT & PATH vs GraphPath & -0.002 [-0.064, +0.060] & 0.945 \\
Prostate & TMT & GraphPath vs BINN & -0.033 [-0.076, +0.008] & 0.114 \\
Prostate & RT & PATH vs BINN & -0.030 [-0.083, +0.024] & 0.271 \\
Prostate & RT & PATH vs GraphPath & +0.008 [-0.074, +0.090] & 0.861 \\
Prostate & RT & GraphPath vs BINN & -0.038 [-0.105, +0.030] & 0.266 \\
Prostate & OS & PATH vs BINN & -0.322 [-0.546, -0.096] & 0.008$^{\dagger}$ \\
Prostate & OS & PATH vs GraphPath & -0.195 [-0.452, +0.050] & 0.125$^{\dagger}$ \\
Prostate & OS & GraphPath vs BINN & -0.127 [-0.292, +0.042] & 0.142$^{\dagger}$ \\
\midrule
Head \& Neck & TMT & PATH vs BINN & -0.052 [-0.148, +0.044] & 0.283 \\
Head \& Neck & TMT & PATH vs GraphPath & -0.011 [-0.113, +0.087] & 0.832 \\
Head \& Neck & TMT & GraphPath vs BINN & -0.041 [-0.103, +0.020] & 0.196 \\
Head \& Neck & RT & PATH vs BINN & -0.069 [-0.173, +0.032] & 0.189 \\
Head \& Neck & RT & PATH vs GraphPath & -0.010 [-0.111, +0.091] & 0.846 \\
Head \& Neck & RT & GraphPath vs BINN & \textbf{-0.059 [-0.106, -0.013]} & 0.013 \\
Head \& Neck & OS & PATH vs BINN & -0.163 [-0.367, +0.030] & 0.098$^{\dagger}$ \\
Head \& Neck & OS & PATH vs GraphPath & -0.163 [-0.328, -0.008] & 0.038$^{\dagger}$ \\
Head \& Neck & OS & GraphPath vs BINN & -0.001 [-0.105, +0.092] & 0.975$^{\dagger}$ \\
\midrule
Thyroid & TMT & PATH vs BINN & +0.120 [-0.091, +0.367] & 0.291$^{\dagger}$ \\
Thyroid & TMT & PATH vs GraphPath & +0.184 [-0.167, +0.569] & 0.320$^{\dagger}$ \\
Thyroid & TMT & GraphPath vs BINN & -0.064 [-0.425, +0.213] & 0.760$^{\dagger}$ \\
Thyroid & RT & PATH vs BINN & +0.070 [-0.112, +0.253] & 0.461 \\
Thyroid & RT & PATH vs GraphPath & +0.032 [-0.144, +0.214] & 0.720 \\
Thyroid & RT & GraphPath vs BINN & +0.038 [-0.076, +0.152] & 0.526 \\
Thyroid & OS & PATH vs BINN & +0.593 [+0.462, +0.722] & <0.001$^{\dagger}$ \\
Thyroid & OS & PATH vs GraphPath & +0.704 [+0.574, +0.815] & <0.001$^{\dagger}$ \\
Thyroid & OS & GraphPath vs BINN & -0.111 [-0.208, -0.019] & 0.034$^{\dagger}$ \\
\bottomrule
\end{tabular}%
}
\end{table}

\section{Training and Reproducibility}\label{app:train}

\paragraph{Reproducibility and availability:}
All models are implemented in PyTorch 2.0 ~\cite{paszke2019pytorch}, with scikit-learn ~\cite{pedregosa2011sklearn} used for
metric computation. Random seeds are fixed identically across Python,
NumPy, and PyTorch for all models and cohorts; the repeated-split sweep is
driven by \texttt{slurm/20\_cv.sbatch} and aggregated by
\texttt{paper/scripts/aggregate\_cv.py}. The single-seed figures are
regenerated by executing \texttt{scripts/run\_all.sh} within the
\texttt{binn/}, \texttt{graphpath/}, and \texttt{path/} directories,
followed by \texttt{paper/build.sh}. Source code, the exact per-cohort,
per-seed split index files, and the per-sample test predictions used for the
bootstrap tests are released so that every reported number and split can be
reproduced exactly.

\end{document}